\documentclass[10pt,twocolumn,letterpaper]{article}

\usepackage{iccv}
\usepackage{times}
\usepackage{epsfig}
\usepackage{graphicx}
\usepackage{amsmath}
\usepackage{amssymb}
\usepackage{enumitem}

\usepackage{bbm} 
\usepackage[ruled,vlined,linesnumbered]{algorithm2e} 
\usepackage{mathtools} 
\usepackage{amsmath}
\usepackage{subcaption}
\usepackage{multirow}
\usepackage{comment}
\usepackage{nicefrac}

\newcommand{\W}{\textbf{\textsf{W}}\xspace}
\newcommand{\X}{\textbf{\textsf{X}}\xspace}
\newcommand{\Y}{\textbf{\textsf{Y}}\xspace}
\newcommand{\XS}{\mathcal{X}\xspace}
\newcommand{\YS}{\mathcal{Y}\xspace}
\newcommand{\WS}{\mathcal{W}\xspace}

\newcommand{\SKf}{\sigma_{\mathrm{f}}} 
\newcommand{\SKc}{\sigma_{\mathrm{c}}} 

\newcommand{\R}{\mathbb{R}\xspace}

\newcommand{\II}{\mathcal{I}_{\text{in}}\xspace}
\newcommand{\IO}{\mathcal{I}_{\text{out}}\xspace}
\newcommand{\G}{\mathcal{G}\xspace}
\newcommand{\GG}{G}

\newcommand{\HH}{H}
\newcommand{\WW}{W}
\newcommand{\KH}{K_{\text{h}}}
\newcommand{\KW}{K_{\text{w}}}
\newcommand{\CI}{C_{\text{in}}}
\newcommand{\CO}{C_{\text{out}}}

\newcommand{\norm}[1]{\left\lVert#1\right\rVert}
\newcommand{\CR}{\mathcal{C}}

\newcommand{\Npar}{N_{\text{par}}}
\newcommand{\Nops}{N_{\text{ops}}}
\newcommand{\Nparmax}{N_{\text{par}}^\text{max}}
\newcommand{\Nopsmax}{N_{\text{ops}}^\text{max}}

\newcommand{\OO}{\mathcal{O}} 

\DeclareMathOperator*{\diag}{\mathsf{diag}}

\DeclareMathOperator*{\argmin}{arg\,min}

\newcommand{\figref}[1]{Figure~\ref{#1}}

\usepackage[pagebackref=true,breaklinks=true,letterpaper=true,colorlinks,bookmarks=false]{hyperref}

\iccvfinalcopy 

\def\iccvPaperID{1965} 

\ificcvfinal\fi
\begin{document}

\title{Efficient Structured Pruning and Architecture Searching for Group Convolution}

\author{Ruizhe Zhao\quad Wayne Luk\\
Imperial College London\\
{\tt\small \{ruizhe.zhao15,w.luk\}@imperial.ac.uk}
}

\maketitle

\begin{abstract}
Efficient inference of Convolutional Neural Networks is a thriving topic recently. It is desirable to achieve the maximal test accuracy under given inference budget constraints when deploying a pre-trained model. Network pruning is a commonly used technique while it may produce irregular sparse models that can hardly gain actual speed-up. Group convolution is a promising pruning target due to its regular structure; however, incorporating such structure into the pruning procedure is challenging. It is because structural constraints are hard to describe and can make pruning intractable to solve. The need for configuring group convolution architecture, i.e., the number of groups, that maximises test accuracy also increases difficulty.

This paper presents an efficient method to address this challenge. We formulate group convolution pruning as finding the optimal channel permutation to impose structural constraints and solve it efficiently by heuristics. We also apply local search to exploring group configuration based on estimated pruning cost to maximise test accuracy. Compared to prior work, results show that our method produces competitive group convolution models for various tasks within a shorter pruning period and enables rapid group configuration exploration subject to inference budget constraints.
\end{abstract}
\section{Introduction}\label{sec:intro}

Convolutional Neural Networks (CNNs) are being deployed on devices ranging from large servers to small edge systems that have various computing capability.
While deploying pre-trained CNN models, we intend to maximise their test accuracy under inference budget constraints, e.g., maximum numbers of parameters and operations.
Pruning is a workable solution that removes parameters contributing little to test accuracy, and its success has been demonstrated in numerous prior works~\cite{Han2015, Han2016, Molchanov2017, Liu2017}.
However, many pruned models can hardly achieve practical test-time speed-up due to irregular sparsity, which results in imbalanced workloads that only customised GPU kernels or specialised hardware~\cite{Han2016} can handle.
Therefore, we are motivated to prune pre-trained models into compact, accurate, and regular sparse models.

\begin{figure}[!t]
  \centering
  \includegraphics[width=\columnwidth]{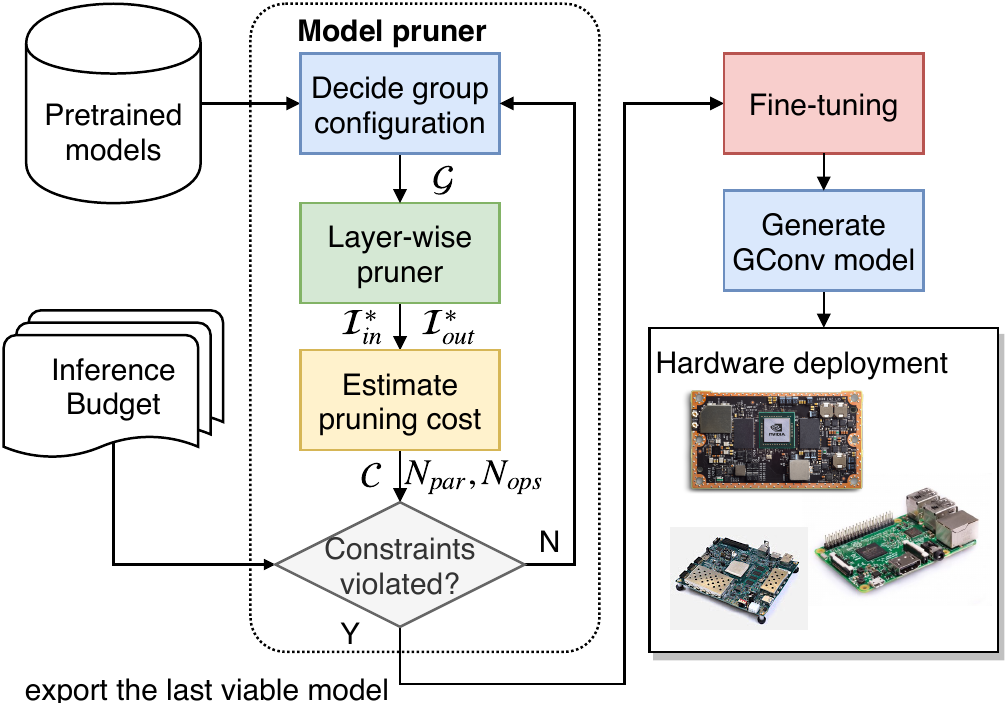}
  \caption{The overall work-flow. Symbols are explained in Section~\ref{sec:method}.}
  \label{fig:workflow}
\end{figure}

\emph{Group convolution} (GConv)~\cite{Krizhevsky2012,Xie2016,Ioannou2016} is a promising pruning target.
A GConv layer consists of multiple identically configured convolution layers, which as a whole can be considered as a \emph{regular} sparse convolution layer with equivalent sparsity across its channels.
GConv also has good learning capability as presented in~\cite{Xie2016,Ioannou2016,Zhang2017a}.
Given these potential benefits,
we expect that pruning pre-trained CNNs into GConv-based models can improve test-time performance regarding speed and accuracy.

However, pruning into GConv is a challenging \emph{structured pruning} problem, i.e., pruned parameters should follow patterns of their positions on input and output channel axes.
These structural constraints turn pruning into a hard-to-solve combinatorial optimisation problem.
Meanwhile, the number of groups should be determined for all layers, which is not a trivial procedure as well.
These two problems have not been properly resolved by prior works~\cite{Zhao2018a,Peng2018,Huang2017a} yet:
they may require training from scratch, manually determining group configuration, or adding overhead during inference.
There is still room for improvement.

\begin{figure*}[t]
  \centering
  \includegraphics[width=0.9\textwidth]{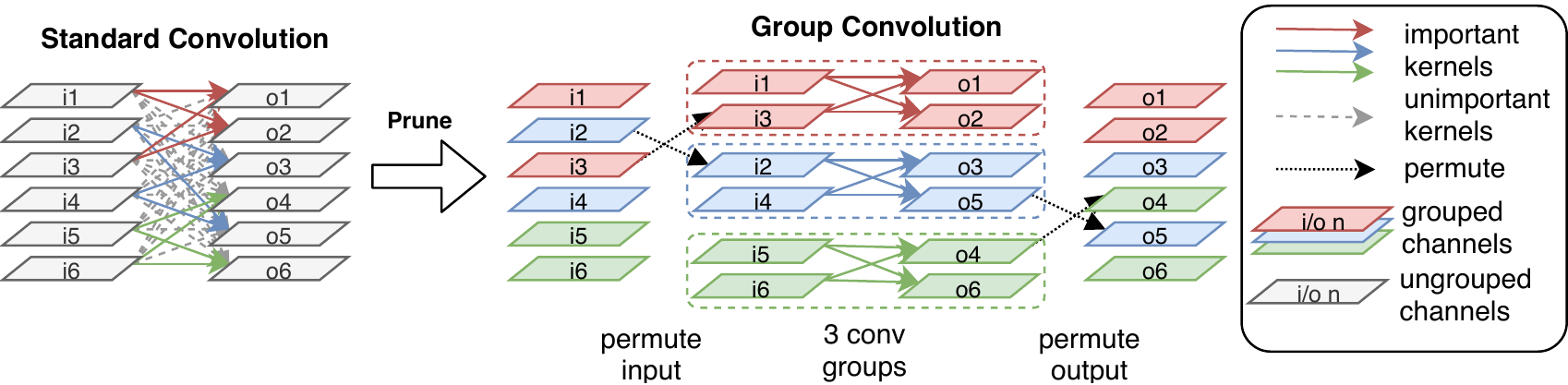}
  \caption{
    An overview of our pruning objective.
    The convolution layer has 6 input and output channels and we intend to prune it into a group convolution with 3 groups.
    Our method will figure out which channels should belong to which groups. 
    The measurement of kernel importance is explained in Section~\ref{sec:method}.
  }
  \label{fig:prune}
\end{figure*}


This paper presents a novel GConv pruning method that addresses both challenges.
For the structured pruning problem, we formulate it as finding the optimal channel permutation that implicitly imposes the structural constraints of GConv and solve it rapidly through heuristics. This solution is named as \emph{layer-wise pruner} (Section~\ref{sec:alg}).
To determine the number of groups for each layer, we employ the layer-wise pruner to estimate the cost of pruning with different given numbers of groups, and apply \emph{local search} to explore feasible solutions within limited time.
This part is introduced as model pruner (Section~\ref{sec:model}).
Finally, we prune the model by the best sparsity configuration that has been explored.
We follow the spirit in~\cite{Liu2019} to either fine-tune the pruned model or train from scratch the topology.
Empirically compared to prior papers, our method produces GConv models that run efficiently in test-time, requires shorter pruning period, and further allows the exploration of sparsity configurations subject to inference budget constraints (Section~\ref{sec:results}).
Our code-base is publicly available\footnote{\url{https://anonymous.4open.science/r/107e12dd-7716-4e1b-81c9-2ea97ae5544a/}}.




\section{Background and Related Work}\label{sec:background}


\paragraph{Group Convolution.}

%
A GConv layer works by partitioning its input channels into disjoint groups and separately convolving each with a group-specific set of filters.
Concretely, given an input tensor shaped $(\CI,\HH,\WW)$, we run $\GG$ convolution layers between each pair of $(\CI/\GG,\HH,\WW)$ input partition and $(\CO/\GG,\CI/\GG,\KH,\KW)$ weight group.
$G$ denotes the number of groups and indicates the sparsity of the GConv layer.
A GConv is sparser when $G$ is larger.
$\CO$ is the number of output channels and $(\KH, \KW)$ is the kernel shape.
Output from these convolution layers are concatenated along the channel axis to produce the final result.

\begin{figure}[t]
  \centering
  \begin{subfigure}[t]{0.35\columnwidth}
    \centering
    \includegraphics[width=\textwidth]{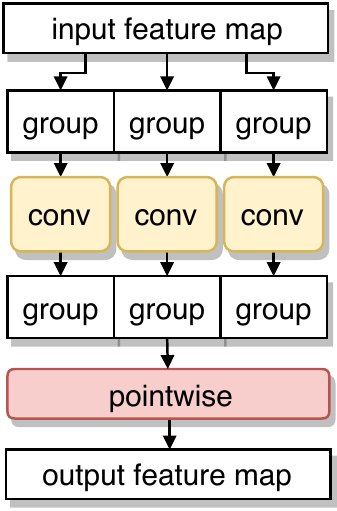}
    \caption{Pointwise-based}
    \label{fig:gconv:pointwise}
  \end{subfigure}
  ~
  \begin{subfigure}[t]{0.35\columnwidth}
    \centering
    \includegraphics[width=\textwidth]{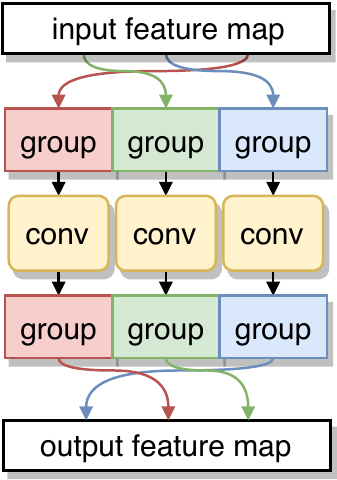}
    \caption{Permutation-based}
    \label{fig:gconv:perm}
  \end{subfigure}
  \caption{\small Two variants of GConv that are different in information exchange mechanism.}
  \label{fig:gconv}
\end{figure}

To improve the learning capacity, we need to encourage information exchange among groups~\cite{Zhang2017a}.
%
\cite{Xie2016,Ioannou2016,Peng2018} suggest using \emph{pointwise} convolution, which applies $1\times 1$ kernels to correlate channels (\figref{fig:gconv:pointwise}).
It is versatile while it can incur unbearable overhead: $\CI\CO$ additional parameters and $\CI\CO H W$ more FMA\footnotemark{} operations are required.
\footnotetext{FMA stands for \emph{fused multiply-add} operation.}
Additionally, it cannot deal with $1\times 1$ group convolution, which is critical since recent efficient CNN heavily rely on them~\cite{Howard2017, Sandler2018, Zhang2017a}.
\cite{Zhao2018a} applies block Hadamard transform, which is more efficient but still requires extra computation.
On the other hand, permuting channels is a much simpler way to mingle groups (\figref{fig:gconv:perm}) since neither additional FMA nor parameter is required.
\cite{Zhang2017a,Zhang2017d,Xie2018a} permute by interleaving channels from different groups, which is also called \emph{channel shuffle}.
We apply permutation as well for its efficiency and the optimisation purpose (Section~\ref{sec:method}).

To construct a CNN by GConv, one can build and train from scratch~\cite{Zhang2017a,Ioannou2016,Xie2016,Zhao2018a,Xie2018}, or prune from pre-trained models.
CondenseNet~\cite{Huang2017a} prunes by a multi-stage, from-scratch training and regularisation procedure.
FLGC~\cite{Wang2019a} follows a similar approach to optimise the GConv topology while training from scratch.
Peng et al.~\cite{Peng2018} consider a GConv layer as a low-rank approximation of a convolution layer.
This approach produces models with high test accuracy but always needs to add pointwise layers.

\paragraph{Network Pruning.}

Our method can be regarded as structural, sensitivity-based network pruning.
Sensitivity pruning selects weights that contribute little to test accuracy and removes them directly based on specific \textit{criteria}, including:
magnitude, e.g., L1, L2 norms~\cite{Mao2017, Han2016, Li2017a, Luo2017},
first-order~\cite{Molchanov2017, Figurnov2015} or
second-order~\cite{LeCun1990, Hassibi1993a, Dong2017} gradients,
average percentage of zero~\cite{Luo2017, Hu2016},
singular values~\cite{Peng2018, Masana2017}.
\cite{Yu2018} considers the importance as a global score.
Each criterion has different computation efficiency and measurement accuracy on contribution from weights.

Alternatively, there are regularisation based methods that sparsify models through curated regularisers so that models can have enough time to adapt.
L1 norm is studied for sparsifying CNN models in \cite{Liu2017, Han2016}, and it normally produces unstructured, irregular models.
Some other methods use \emph{group LASSO} \cite{Yuan2006} to encode specific structures during regularisation, such as channel or filter level pruning \cite{Pan2016, Wen2016, Lebedev2016}.
Specifically, CondenseNet~\cite{Huang2017a} adapts this method to GConv pruning.
Pruning by LASSO regularisers is more difficult due to non-differentiability around optimal points and hyperparameters are hard to tune.

\paragraph{Neural Architecture Search (NAS).}
NAS is a recently developed technique that enables automatic exploration of neural network architectures under specific constraints.
The core mechanism behind is normally the REINFORCE algorithm~\cite{Williams1992}, specifically, \cite{Zoph2016, Zoph2017, Tan2018, Tan2019} consider searching architectures under different platform constraints.
Evolutionary algorithm is another option~\cite{Real2018,Chen2019,Liu2017a}.
To make NAS more efficient, using gradient-based method~\cite{Liu2018c} or reducing the search space~\cite{Liu2018a} is proposed as well.
Parts of our method overlap the objective of NAS: we intend to search for the group configuration under inference budget.
This is a novel objective and we provide an efficient solution based on our pruning method, and we are intrigued to see how mainstream NAS algorithms can be applied to this research question.
\section{Method}\label{sec:method}

Our objective is to \emph{prune} a pre-trained CNN model into a GConv based model (\figref{fig:prune}).
The following sections formulate GConv pruning as an optimisation problem that searches for channel permutations, demonstrate the layer-wise heuristic pruning algorithm that efficiently solves this problem, and show how to explore model sparsity under inference budget constraints.

\subsection{Group Convolution Pruning}

\begin{equation}\label{eq:gconv}
  \begin{aligned}
    & \tilde{\X}_c = \X_{\II(c)} \quad
      \tilde{\Y}^{g} = \W^{g} * \tilde{\X}^{g}\quad
      \Y_f = \tilde{\Y}_{\IO(f)} \\
    & \forall\ 1 \leq f \leq \CO \quad
               1 \leq c \leq \CI \quad
               1\leq g\leq G
  \end{aligned}
\end{equation}

\paragraph{Group convolution.}

We can formulate GConv by mapping from \figref{fig:gconv:perm} to~\eqref{eq:gconv}.
$\X$ and $\Y$ are both 3D \emph{tensors} indicating input and output feature maps;
$\X_c$ and $\Y_f$ are 2D images belong to input channel $c$ and output channel $f$ respectively;
and $\W\in\R^{\CO\times\CI\times\KH\times\KW}$ is a 4D tensor denoting weights.
$\II$ and $\IO$ are two sets of channel indices that specify permutation, e.g., $\II(c)$ is permuted to $c$.
When performing GConv, we permute $\X$ into $\tilde{\X}$ by input channel indices $\II$,
partition $\tilde{\X}$ into $\GG$ groups $\{\tilde{\X}^g\in\R^{\CI/G\times\HH\times\WW}\}_{g=1}^\GG$ along channel,
run a convolution layer ($*$) between $\X^g$ and weight group $\W^g\in\R^{\CO/G\times\CI/G\times\KH\times\KW}$ for each group $g$,
concatenate their output $\{\tilde{\Y}^g\}_{g=1}^G$ into a tensor $\tilde{\Y}$ by channel,
and permute $\tilde{\Y}$ by output channel indices $\IO$ to produce the result $\Y$.

Alternatively, we can treat GConv as a sparse convolution.
To illustrate this idea, we reshape $\W$ into a 6D tensor $\WS\in\R^{\GG\times\GG\times\CO/\GG\times\CI/\GG\times\KH\times\KW}$ by partitioning the input and output channels of $\W$ into $\GG$ groups.
Considering each $\KH\times\KW$ kernel as a single element, this representation can be viewed as a generalised \emph{block matrix} with $\GG\times\GG$ number of $\CO/\GG\times\CI/\GG$ sized blocks.
Similarly, we can partition $\X$ and $\Y$ into 4D tensors $\XS\in\R^{\GG\times\CI/\GG\times\HH\times\WW}$ and $\YS\in\R^{\GG\times\CO/\GG\times\HH\times\WW}$ and view them as generalised \emph{block vectors}.
Note that 2D images are considered as elements in these block matrices.

\begin{align}\label{eq:blockconv}
  &\YS_{g_f}
      = \sum_{g_c=1}^G \WS_{g_f g_c} * \XS_{g_c}
      \quad \forall\ 1\leq g_f\leq G \\
  & \begin{aligned}
    \YS &= \diag(\{\WS_{gg}\}_{g=1}^G) * \XS          \\
        &= \begin{bmatrix}
              \WS_{11} & 0        & \cdots & 0       \\
              0        & \WS_{22} & \cdots & 0       \\
              \vdots   & \vdots   & \ddots & \vdots  \\
              0        & 0        & \cdots & \WS_{GG} 
            \end{bmatrix}
            \begin{bmatrix}
            \XS_1 \\
            \XS_2 \\
            \vdots \\
            \XS_G \\
            \end{bmatrix}
            = 
            \begin{bmatrix}
            \XS_1 * \WS_{11} \\
            \XS_2 * \WS_{22} \\
            \vdots    \\
            \XS_G * \WS_{GG} \\
            \end{bmatrix}\label{eq:blockmat}
  \end{aligned}
\end{align}

We can define convolution among group partitioned tensors $\YS=\WS*\XS$ as a generalised \emph{block matrix-vector multiplication}.
Here, the multiplication between entries in $\WS$ and $\XS$ is interpreted as convolution.
For example, \eqref{eq:blockconv} illustrates the dot-product routine that computes the convolution between two blocks in $\WS$ and $\XS$.
Interestingly, if $\WS$ is a generalised \emph{block-diagonal} matrix, i.e, only $\WS_{gg}$ are non-zero, then this convolution becomes a group convolution~\eqref{eq:blockmat}.
Ignoring permutations in~\eqref{eq:gconv} and considering $\WS_{gg}$ as $\W^g$, $\XS_g$ as $\X^g$, and $\YS_g$ as $\Y^g$, it is obvious that~\eqref{eq:gconv} is equivalent to~\eqref{eq:blockmat}.
This is the basis of the following analysis.


\begin{equation}\label{eq:perm_exp}
\begin{bmatrix}
  w_{11} & w_{12} & w_{13} \\
  w_{21} & w_{22} & w_{23} \\
  w_{31} & w_{32} & w_{33}
\end{bmatrix}
\xrightarrow[\II=\{1, 3, 2\}]{\IO=\{2, 1, 3\}}
\begin{bmatrix}
  w_{21} & w_{23} & w_{22} \\
  w_{11} & w_{13} & w_{12} \\
  w_{31} & w_{33} & w_{32}
\end{bmatrix}
\end{equation}

\paragraph{Pruning and channel permutation.}

Pruning means removing weights unimportant to model accuracy from a trained model based on a specific criterion.
\eqref{eq:blockmat} shows that removing weights to form a block-diagonal matrix is equivalent to pruning into GConv.
A straightforward approach to prune is just removing kernels outside the diagonal.
It can hardly perform well since we cannot guarantee that kernels around the diagonal are important to the model accuracy.
Since we allow channel permutation on both input and output, we can formulate pruning as an optimisation problem that targets at \emph{finding the channel permutation that can move most of the important kernels to diagonal blocks}.
To be specific on permutation, \eqref{eq:perm_exp} shows an example result after applying a pair of permutation indices $\IO$ and $\II$ on weights.
$w_{ij}$ denotes a single kernel, and rows and columns represent output and input channel axes respectively.

%

\paragraph{Optimisation problem.}
\eqref{eq:opt} formulates the optimisation problem of finding optimal permutations $\II^*$ and $\IO^*$.
We need to find the pair of permutations such that, after applying it on the original weights $\W$, the importance reduction caused after removing weights outside the diagonal will be minimal.
$\{\tilde{\WS}_{gg}\}_{g=1}^G$ denotes all the diagonal-blocks of permuted weights.
$\CR$ is the criterion that measures the importance (Section~\ref{sec:background}).
We choose a magnitude-base criterion that sums the L2 norm of all kernels, based on the assumption that kernels with greater magnitude contribute more to model accuracy.

\begin{equation}\label{eq:opt}
\begin{aligned}
  \argmin_{\II^*, \IO^*}
    &\quad \CR(\W) - \CR\left(\diag(\{\tilde{\WS}_{gg}\}_{g=1}^G)\right) \\
  \mathbf{s.t.}
    &\quad \tilde{\WS}_{g_f g_c} \equiv \tilde{\W}^{g_f g_c}
     \quad \tilde{\W}_{fc}  = \W_{\IO^*(f)\II^*(c)}
\end{aligned}
\end{equation}

We notice that solving this problem requires similar efforts as solving the Bottleneck Travelling Salesman Problem (BTSP)~\cite{Garfinkel1978}, which is known to be NP-complete.
Since the number of channels can be hundreds or even thousands, directly solving this problem is computationally intractable.
Next section presents a heuristic algorithm that produces satisfiable solutions within a limited time.


%
%

\subsection{Prune a Layer}\label{sec:alg}

This section focuses on the layer-wise pruning problem: for a given sparsity, indicated by the number of groups $\GG$, we aim to find the pair of column-wise and row-wise permutations that minimise the pruning objective defined in~\eqref{eq:opt}.
Column and row refer to input and output channel axes of weights respectively.

\begin{equation}\label{eq:max}
\begin{aligned}
  & \max_{\IO,\II} \quad
    \sum_{g=1}^G 
    \sum_{f=1}^{\nicefrac{\CO}{\GG}}
    \sum_{c=1}^{\nicefrac{\CI}{\GG}}
    \norm{\tilde{\WS}_{g g f c}}
\end{aligned}
\end{equation}

We start from replacing $\CR$ with L2 norm to convert the original problem to an equivalent maximisation problem that \emph{maximises the sum of L2 norm of weight kernels in diagonal blocks} of $\tilde{\WS}$, as shown in~\eqref{eq:max}.

\begin{equation}\label{eq:sort_keys}
\begin{aligned}
\SKc(c,g) &= \sum\nolimits_{f=\frac{g-1}{\GG}\CO+1}^{\frac{g}{\GG}\CO} \norm{\W_{f c}}\\
\SKf(f,g) &= \sum\nolimits_{c=\frac{g-1}{\GG}\CI+1}^{\frac{g}{\GG}\CI} \norm{\W_{f c}}
\end{aligned}
\end{equation}

\paragraph{Sub-problem.}
Suppose we only maximise the sum-of-norm for the $G$-th block diagonal component $\WS_{G,G}$, and we are only allowed to permute input channels, an intuitive solution for this sub-problem is sorting input channels by $\SKc(c,G)$~\eqref{eq:sort_keys}.
Under the given restrictions, sorting by $\SKc$ in an increasing order practically moves the most important weights to $\WS_{G G}$.
Similarly, if only output channels are permitted to permute, we can sort them by $\SKf(f,G)$  as well to maximise the importance of $\WS_{G G}$.

\paragraph{Heuristic Algorithm.}
This intuition leads us to a heuristic algorithm that solves~\eqref{eq:max}.
Instead of considering this problem as a whole, we dissect it into sub-problems similar to the example above, which can be solved block by block through sorting with regards to $\SKc$ and $\SKf$.
Specifically: 

\begin{enumerate}[itemsep=0em,label=(\roman*)]
\item Our algorithm runs in $G$ iterations and the $g$-th iteration works on optimising block $\WS_{G-g+1,G-g+1}$ only.
\item Each iteration sorts input and output channels by $\SKc(c,g)$ and $\SKf(f,g)$ respectively for $N_S$ rounds.
\item All channels related to previously resolved blocks remain \emph{frozen}, i.e., the upper bounds for $f$ and $c$ that can be sorted are $(\GG-g+1)\CO/\GG$ and $(\GG-g+1)\CI/\GG$.
\end{enumerate}

\begin{figure}[!t]
\centering
\includegraphics[width=0.8\columnwidth]{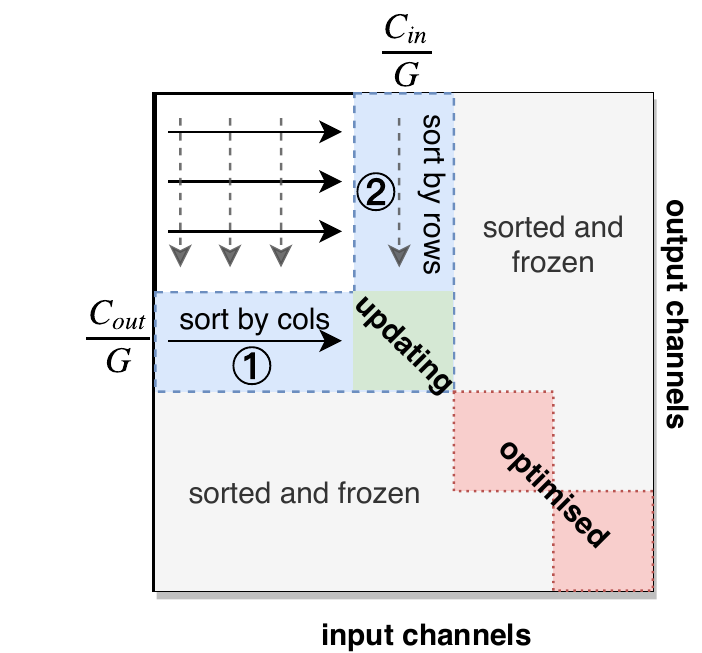}
\caption{
\small
This figure shows $\WS$ that is being sorted.
It is running at the 3rd block (green).
Kernels covered by previous blocks (red) are now frozen for further sorting (gray).
When updating the 3rd block, input channels (columns) are sorted by $\SKc(c,3)$ in range \textcircled{1} and output channels (rows) are sorted by $\SKf(f,3)$ in range \textcircled{2}.}
\label{fig:alg}
\end{figure}

\figref{fig:alg} illustrates an intermediate step of this algorithm.
$N_S$ is a hyperparameter that denotes the number of sorting rounds for each block.
A sorting round means sorting input channels and then output channels.
$N_S$ is necessary since we permit sorting both input and output channels: after finishing a sorting round, the increasing order of input channels regarding $\SKc$ may be violated, and running a new round may fix it.
For example, as shown in \figref{fig:alg}, kernels in the white region are not covered in $\SKc$ when sorting input channels at first, and after output channels are sorted, some may enter the blue region and affect the evaluation of $\SKc$.
What we present here is a polynomial time algorithm and its complexity is
$\OO(\GG N_S \times (\frac{\CI}{\GG}\log(\frac{\CI}{\GG}) + \frac{\CO}{\GG}\log(\frac{\CO}{\GG})))$.

\begin{equation}\label{eq:ratio}
  \text{recovery ratio} =
  \frac{\sum_{g,f,c} \norm{\tilde{\WS}_{g g f c}^*}}{\sum_{g,f,c} \norm{\WS_{g g f c}}}
\end{equation}

\paragraph{Empirical Evaluation.}
We empirically justify this algorithm by running it on randomly generate sample weights and real-world pre-trained models.
We first create a block-diagonal matrix, then permute it by random indices, and try to recover the original permutation as much as possible.
We measure the quality by the \emph{recovery ratio} defined by~\eqref{eq:ratio}, in which $\tilde{\WS}^*$ denotes weights permuted by optimal permutations decided by our algorithm.
We notice that the recovery ratio becomes higher for more samples when $N_S$ increases, and when $N_S=10$, most samples can achieve 100\% recovery ratio.
It shows that by using our heuristic algorithm with $N_S=10$ we can move most important weights to diagonal blocks, which implicitly guarantees our GConv pruning performance since we remove weights out of diagonal blocks.

\begin{figure}[!t]
  \centering
  \includegraphics[width=\columnwidth]{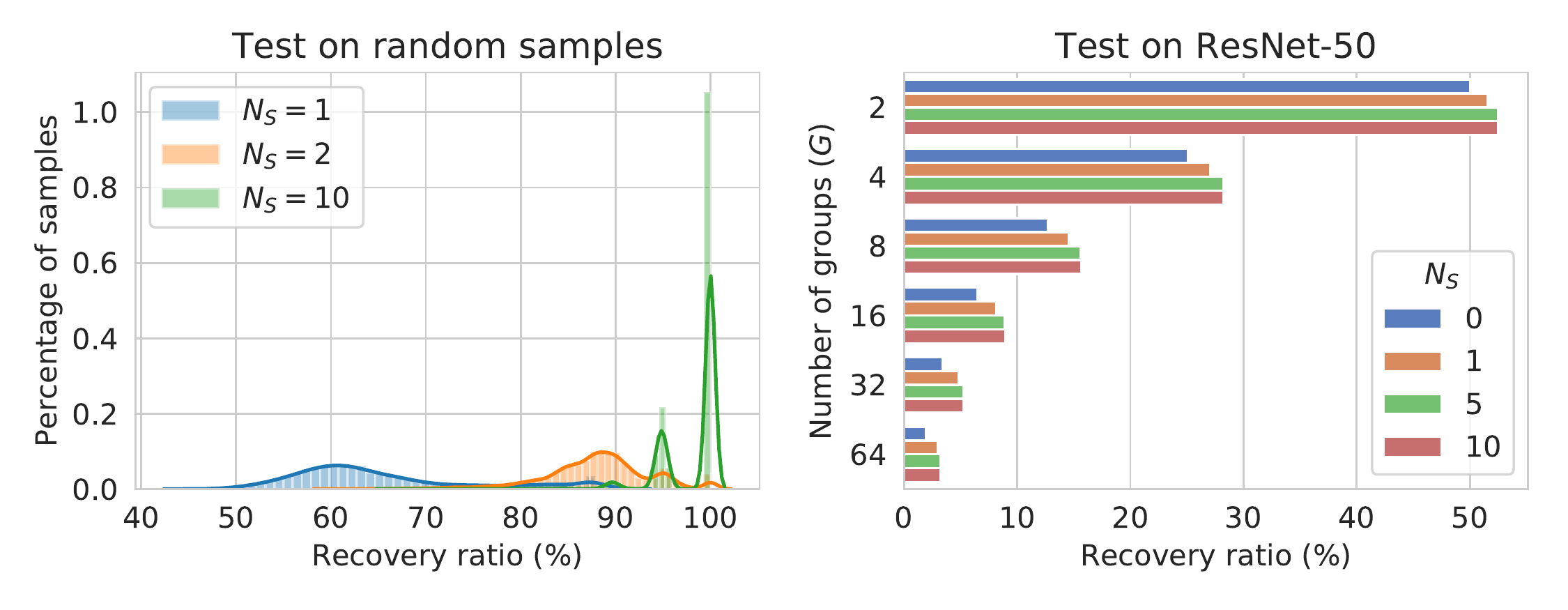}
  \caption{\small
  Evaluation results on the heuristic algorithm. Recovery ratio is the percentage of magnitude, measured by L2 norm, retained in the diagonal blocks after pruning.}
  \label{fig:perf}
\end{figure}

We also measure the recovery ratio on weights from pre-trained ResNet-50\footnotemark{}:
as shown on the right of~\figref{fig:perf}, compared with the baseline that does no sorting ($N_S=0$), our method can recover about 3\% more for different numbers of groups.
Besides the recovery ratio, we compare the final test accuracy between using and not using our heuristic method in Table~\ref{table:ablation:perm}.
In the future, we will provide formal proof regarding the performance of this heuristic algorithm.

\footnotetext{Pre-trained models are downloaded from \url{https://pytorch.org/docs/stable/torchvision/models.html}}

\begin{table*}[!t]
  \small
  \centering
  \caption{
    Evaluation on CIFAR-10 (C10) and CIFAR-100 (C100).
    Number of parameters (\# Param.) and operations (\# FLOPS) are measured on CIFAR-100.
    Note that we consider one FMA as two operations.
    The DenseNet-86 on the last line is named as CondenseNet-86 in its original paper~\cite{Huang2017a}.}
  \label{table:CIFAR}
  \begin{tabular}{l|cc|cccc}
    \hline
     & \multicolumn{2}{c|}{\bf Test Error (\%)} & & & & \\
    \cline{2-3}
    {\bf Model} & C10 & C100 &
    {\bf \# Param.} & {\bf Pruned} & {\bf \# FLOPS} & {\bf Pruned} \\
    \hline
    ResNet-164 (Baseline)
      & 4.81 & 22.83 & 1.73M & ---      & 504M & ---     \\
    ResNet-164 (40\% pruned)
      & 4.91 & 22.71 & 1.46M & 15.5\%   & 462M & 8.4\% \\
    ResNet-164 (60\% pruned) 
      & 5.08 & 23.66 & 1.21M & 29.7\%   & 430M & 14.7\% \\
    \hline
    ResNet-164 (40\% pruned)~\cite{Liu2017}
      & 5.08 & 22.87 & 1.46M & 15.5\% & 333M & 33.3\% \\
    ResNet-164 (60\% pruned)~\cite{Liu2017}
      & 5.27 & 23.91 & 1.21M & 29.7\% & 247M & 50.6\% \\
    \hline
    \hline
    DenseNet-86 (Baseline)~\cite{Huang2017a}
      & 4.44 & 20.57 & 2.03M & ---     & 506.1M & ---      \\
    DenseNet-86 ($G=4$)
      & 5.94 & 25.96 & 0.59M & 70.94\% & 132.7M & 73.77\% \\
    DenseNet-86 (opt. for 50\% budget)
      & 4.98 & 22.41 & 1.00M & 50.74\% & 256.5M & 49.33\%  \\
    \hline
    CondenseNet-86 ($G=4$)~\cite{Huang2017a}
      & 5.00 & 23.64 & 0.59M & 70.94\% & 132.7M & 73.77\%  \\
    \hline
  \end{tabular}
\end{table*}

\subsection{Optimise Group Configuration}\label{sec:model}


To prune a whole CNN model into one that uses GConv, we need to determine the \emph{group configuration}, which is a combination of all layers' numbers of groups.
Group configuration can affect both the model accuracy and the inference budget, and therefore, finding an optimal group configuration is an optimisation problem that maximises accuracy under budget constraints.
Prior papers use some ad-hoc rules to decide group configuration, e.g., using a uniform group number for all layers~\cite{Huang2017a} or scaling the number of groups by the number of channels~\cite{Peng2018,Zhao2018a}.
When using these methods, the only way to find the configuration that gives the highest model accuracy is by trial-and-error, i.e., manually picking a group configuration and fine-tuning from it, which is cumbersome.
To address this problem, we propose a group configuration optimisation algorithm, which finds near-optimal configuration regarding model accuracy by utilising pre-trained weights through our GConv pruning algorithm.

\begin{equation}\label{eq:gopt}
\begin{aligned}
\argmin_{\G}
  &\
  \sum_{l=1}^L \mathsf{cost}(\WS^{(l)}, \G_l)
  \\
\mathbf{s.t.}
  &\ 
  \sum_{l=1}^L \frac{\Npar^{(l)}}{\G_l} \leq \Nparmax \quad
  \sum_{l=1}^L \frac{\Nops^{(l)}}{\G_l} \leq \Nopsmax \\
\end{aligned}
\end{equation}

This method is formulated as~\eqref{eq:gopt}.
The variable $\G$ is a vector that specifies the number of groups of each layer $l$.
The function $\mathsf{cost}$ denotes the minimal pruning cost returned from solving~\eqref{eq:opt}.
We assume that the lower the pruning cost, the higher the accuracy of a pruned model.
The objective is to minimise the sum of estimated pruning cost of all layers subject to constraints on the maximum numbers of parameters $\Nparmax$ and operations $\Nopsmax$.

Our approach is a local search algorithm~\cite{Russell}.
We notice that by adding the number of groups of a layer, the total cost increases and the numbers of parameters and operations decrease.
Based on this observation, we devise this local search algorithm by starting with a $\G$ that sets $G$ to 1 for all layers, and in each of the following iterations, selecting one layer that minimally increases the cost when its number of groups is changed to the nearest larger candidate.
The whole procedure terminates when the resource constraints are satisfied.
To estimate pruning cost more precisely, we can optionally prune and fine-tune the model by the current $\G$ at the end of each iteration.
This algorithm is summarised in Algorithm~\ref{alg:model}.


\begin{algorithm}[h]
\caption{The local search algorithm to solve~\eqref{eq:gopt}}
\label{alg:model}
$\G\leftarrow \text{maximal numbers of groups of all layers}$\;
\While{$\text{budget of } \G \text{ is under constraints}$}{
    $i \leftarrow \text{layer that reduces cost the most}$\;
    $\G(i) \leftarrow \text{next larger number of groups for } i$\;
    (optional) prune model by $\G$ and then fine tune\;
}
\Return $\G$\;
\end{algorithm}

As a final step, based on the optimised $\G$, we prune and fine-tune the given model again to improve its model accuracy as much as possible.
Empirically we find this algorithm works well. As shown in the next section (\figref{fig:result:gopt}), we can explore configurations within given budget and the explored models perform competitively compared to ad-hoc configurations.

%

\section{Experiments}\label{sec:results}

This section presents various experiments to empirically evaluate our method.

\subsection{Experiment Setup}

\paragraph{Datasets and models.}
Our method is evaluated on CIFAR-10/100~\cite{Krizhevsky2009} and ImageNet~\cite{Russakovsky2015}.
\cite{pytorch-classification} is used to build and train CIFAR-10/100 baseline models. 
Regarding ImageNet, we evaluate on ILSVRC2012 and augment data by random cropping and then random horizontal flipping and the validation accuracy is evaluated by center-cropping.

We choose various CNN models for evaluation:
ResNet-110~\cite{He2015}, ResNet-164~\cite{He2016}, and DenseNet-86~\cite{Huang2016a} for CIFAR-10/100; ResNet-18/34/50/101~\cite{He2015} for ImageNet.
Our models are all implemented by PyTorch~\cite{Paszke2017} v1.1.

\paragraph{Pruning and fine-tuning.}
When deciding group configuration $\G$, we may use a uniform $G$ value for all layers, a configuration borrowed from a prior work, or one generated by solving~\eqref{eq:gopt}.
Once $\G$ is settled, we run the heuristic layer-wise pruning algorithm (Section~\ref{sec:alg}) to get channel permutation, which indicates which weights should be pruned and how to permute input and output channels.
The hyperparameter $N_S$ is normally set to $10$ based on \figref{fig:perf}.

After pruning the model, in the fine-tuning phase, we normally choose a relatively small learning rate to train the pruned model for a few more epochs.
The fine-tuning period could be around one third of the original training from scratch time.
We will tune training hyperparameters further in the future to see at least how much workload is required to recover the accuracy of a pruned model.
\subsection{Results}

\paragraph{ResNet-164 on CIFAR.}
We list our results on CIFAR-10 and 100 in Table~\ref{table:CIFAR}.
We first compare ResNet-164 with network slimming~\cite{Liu2017}, a state-of-the-art channel-wise structured pruning method based on regularisation.
Their models are pruned with respect to the percentage of removed channels, which are quite different from our GConv-based results.
To compare our method with them at a similar pruning level, we set the number of parameters of their models as constraints for our group configuration optimisation, and use optimised configurations to prune ResNet-164.
As shown in Table~\ref{table:CIFAR}, our models have much smaller test error than their counterparts with the same number of parameters. 
Considering model topology, our resulting models are also easier to process: convolution layers with arbitrary amount of channels produced by~\cite{Liu2017} may not be friendly to low-level accelerator, while ours are basically GConv, which runs efficiently on modern hardware.

One drawback of our method on this ResNet-164 case is the relatively higher FLOP number, which is caused by the fact that layers closer to the output normally have lower pruning cost but less contribution to FLOP.
Our method tends to give these layers higher pruning priority.
This issue will be mitigated by introducing FLOP into the pruning cost measurement in our future work.

\paragraph{Comparison with CondenseNet.}
CondenseNet~\cite{Huang2017a} introduces a multi-staged, group-lasso regularisation based GConv pruning procedure.
With a given group configuration, this paper provides the state-of-the-art GConv pruning results on variants of DenseNet~\cite{Huang2016}.
For the comparison purpose, we select DenseNet-86, a variant of DenseNet and is pruned to CondenseNet-86 in~\cite{Huang2017a}, as a baseline model.

Since they put more efforts in training and regularisation, it is hard for our post-training pruning method to surpass their level of accuracy.
Table~\ref{table:CIFAR} shows that our $G=4$ result is around 1-2\% worse on CIFAR-10/100 validation accuracy than the CondenseNet counterpart.
This accuracy loss can be explained by the additional regularisation effect introduced while training CondenseNet.

Even though using our method is still beneficial in some scenarios.
CondenseNet can only be trained by a fixed, manually picked group configuration, while we can explore the group configuration from pre-trained DenseNet models.
We can find more accurate models under different inference budget, e.g., in Table~\ref{table:CIFAR} we show a better option under a looser budget of 50\%.
Performing similar exploration in CondenseNet will take much longer time due to the need to train from scratch.

%
%

\begin{figure}[!t]
  \centering
  \includegraphics[width=\columnwidth]{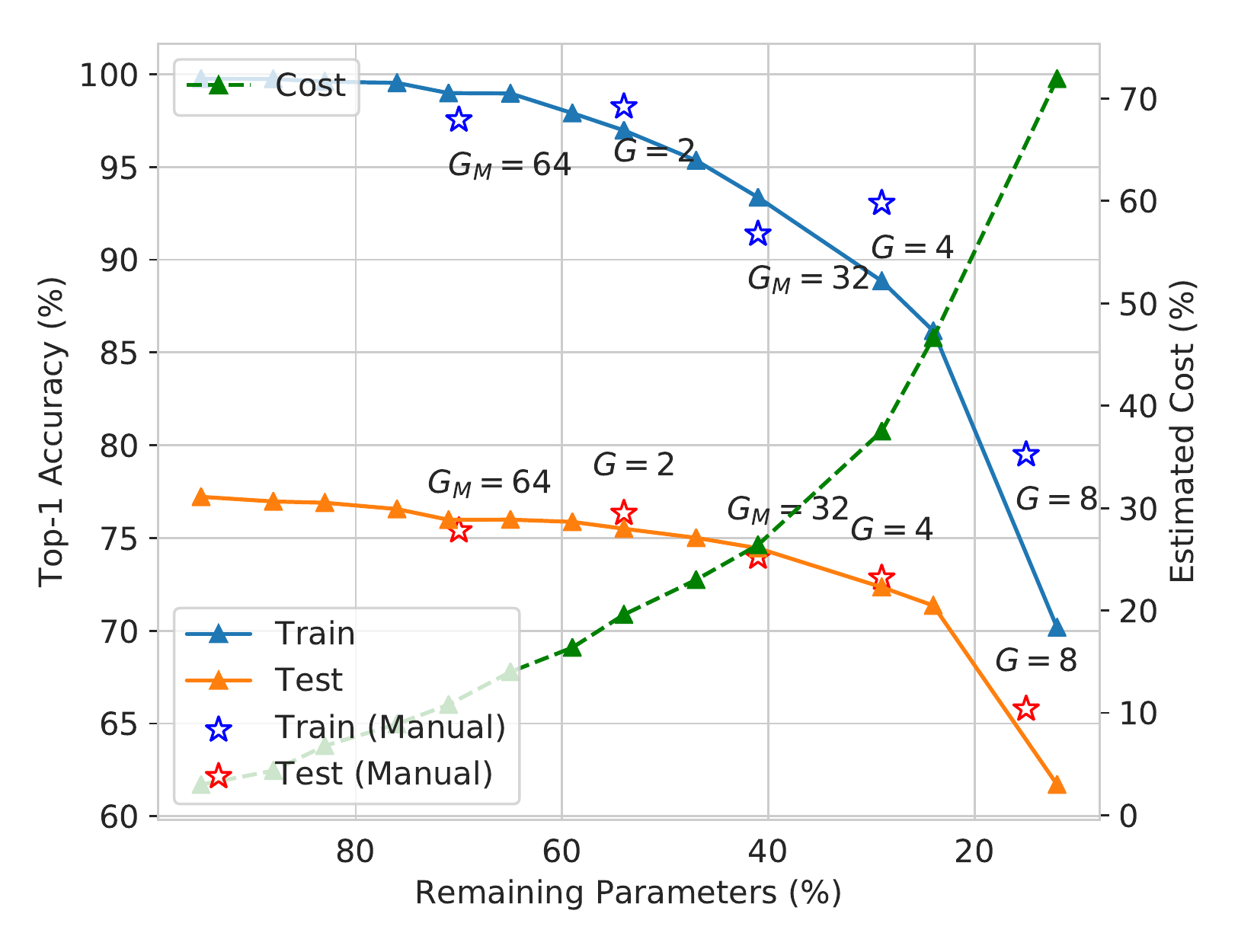}
  \caption{
  \small
  Change in ResNet-164 on CIFAR-100 accuracy while reducing number of parameters.
  Each sample point is collected from the model found by our model pruning algorithm under the given number of parameters constraint.
  Manually configured models either use uniform group size $G$ or set a maximum channel number $G_M$.
  The latter approach assigns each layer a group number that ensures $\max(\CI/\GG,\CO/\GG)\leq\GG_M$.
  We also list the estimated pruning cost for each explored sparsity configuration.
  Accuracy numbers listed here are collected from \emph{fine-tuning}.
  }
  \label{fig:result:gopt}
\end{figure}

\paragraph{Group Configuration Exploration.}
One of our major benefits is that we can explore group configuration under given inference budget constraints, as mentioned earlier.
To evaluate the quality of the exploration, we gradually reduce the upper bound of number of parameters and run the model pruner.
Results for ResNet-164 on CIFAR-100 are in \figref{fig:result:gopt}.
The granularity of exploration here is 0.1M and we use the same training schedule for each sparsity configuration.
Results demonstrate that explored models can perform on par with manually selected configurations.

\paragraph{ImageNet.} 
We evaluate various pre-trained ResNet models on ImageNet~\cite{He2015} .
The fine-tuning phase has 30 epochs (1/3 of what \cite{He2015} uses) with learning rate starting at $1e^{-3}$ and being multiplied by 0.1 every 10 epochs.
Limited by hardware resources, we only select uniform numbers of groups.

\begin{table}[!t]
  \small
  \centering
  \caption{\small ImageNet evaluation results. The first sample in each section is the baseline.}
  \label{table:imagenet}
  \begin{tabular}{l|rrr}
    \hline
    Model                & \# Params. & \# FLOPS & Top-1 Error   \\
    \hline\hline
    ResNet-18            & 11.69M     & 3.64G & 30.24\% \\
    ResNet-18 $G$=8      & 1.91M      & 0.20G & 44.71\% \\
    \hline
    ResNet-18 $G$=8 \cite{Zhao2018a}
                         & 1.91M      & 0.33G & 44.60\% \\
    \hline\hline
    ResNet-34            & 21.80M     & 7.34G & 26.70\% \\
    ResNet-34/A          & 17.36M     & 3.48G & 28.42\% \\
    ResNet-34/B          & 9.05M      & 1.89G & 32.44\% \\
    \hline
    ResNet-34/A\cite{Peng2018}
                         & 18.2M      & 3.98G & 27.05\% \\
    ResNet-34/B\cite{Peng2018}
                         & 11.1M      & 2.62G & 27.75\% \\
    \hline\hline
    ResNet-50            & 25.56M     & 8.21G & 23.85\% \\
    ResNet-50 $G$=2      & 13.82M     & 3.77G & 25.90\% \\
    \hline\hline
    ResNet-101           & 44.55M     & 15.7G & 22.63\% \\
    ResNet-101 $G$=2     & 23.34M     & 7.50G & 24.22\% \\
    \hline
  \end{tabular}
\end{table}

Results are listed in Table~\ref{table:imagenet}.
Since models for ImageNet are rarely sparse, removing many parameters in one-shot normally degrades the accuracy significantly.
For ResNet-18, we can reduce 83\% parameters and 95\% FLOPS with an increase of 14.5\% in test error.
It is not promising regarding the high test error, but this performance is on par with~\cite{Zhao2018a}, even if they have higher budget in both training and inference phases: their ResNet-18 example is trained from scratch and the Hadamard transform adds overhead.

We also compare ResNet-34 to~\cite{Peng2018}, which uses low-rank approximation to perform GConv pruning.
This method requires adding pointwise $1\times 1$ convolution after each GConv.
Since they use more parameters that potentially increase learning capacity, their accuracy can be higher.
For the two configurations from~\cite{Peng2018}, ResNet-34/A and /B, models produced by us are smaller but less accurate.
However, as mentioned before, their GConv is always appended by a pointwise convolution, which means with the same number of groups they use more resources, and they cannot deal with $1\times 1$ group convolution.
Specifically, for ResNet-50/101 that heavily uses $1\times 1$ kernels, our method can still reduce around 50\% budget without increasing much test error.

\subsection{Ablation Study}

This section investigates the effect of different design choices that may appear while using our method.

\begin{table}[!t]
  \small
  \centering
  \caption{\small
    Comparison of the top-1 test error (\%) between using the \emph{heuristic} algorithm and not using (\emph{plain}). ResNet-50 results are collected on ImageNet.}
  \label{table:ablation:perm}
  \begin{tabular}{l|cc|cc}
    \hline
                       & \multicolumn{2}{c|}{heuristic}
                       & \multicolumn{2}{c}{plain} \\
                       \cline{2-5}
    Model              & C10 & C100 & C10 & C100 \\
    \hline
    ResNet-110 ($G=2$) & {\bf 6.69} & {\bf 27.63} & 7.13  & 28.08 \\
    ResNet-110 ($G=8$) & {\bf 9.39} & {\bf 33.28} & 11.17 & 37.29 \\
    \hline
    ResNet-50  ($G=2$) & \multicolumn{2}{c|}{{\bf 25.90}}
                       & \multicolumn{2}{c}{28.04}\\
    \hline
  \end{tabular}
\end{table}

\paragraph{Effect from heuristic algorithm.}

We compare our algorithm with a \emph{plain} algorithm, which tries to perform GConv pruning without sorting channels.
Referring to Section~\ref{sec:alg}, this plain algorithm simply sets $N_S$ to 0.
This comparison is performed under the same group configuration to see whether our heuristic algorithm can improve the resulting accuracy.
Table~\ref{table:ablation:perm} presents the comparison between both approaches on ResNet-110 (CIFAR-10/100) and ResNet-50 (ImageNet) with uniform $G$.
All experiments use the same training scheme (learning rate, number of epochs, etc.).

We notice that by using the heuristic algorithm, the test error of all variants on all datasets is reduced.
This result is also in line with the relative order of recovery ratio as shown in \figref{fig:perf}: $N_S$=0 has less recovery ratio than $N_S$=10 that is used by our heuristic algorithm, and it performs worse regarding the model accuracy.
It can provide a concrete evidence that maximising recovery ratio, or equivalently minimising pruned magnitude, is an effective approach to conduct GConv pruning.

\begin{table}[!t]
  \small
  \centering
  \caption{
    \small
    Comparison between pruning and training from scratch for ResNet-164 on CIFAR-100 regarding top-1 train and test error.}
  \label{table:ablation:scratch}
  \begin{tabular}{l|cc|cc}
    \hline
                & \multicolumn{2}{c|}{Train Err. (\%)}
                & \multicolumn{2}{c}{Test Err. (\%)} \\\cline{2-5}
    Method      & $G$=2 & $G$=4 & $G$=2  & $G$=4 \\
    \hline\hline
    Pruning     & 1.75 & 10.50 & \textbf{23.66} & 27.15  \\ 
    \hline
    Shuffle     & \textbf{0.81} & \textbf{3.55} & 23.77 & \textbf{26.60} \\ 
    None        & 2.58 & 16.28 & 26.63 & 32.00 \\ 
    \hline
  \end{tabular}
\end{table}

\paragraph{Compare with training from scratch.}
Inspired by~\cite{Liu2019}, we try to investigate that besides the reduction in training budget whether pruning can also achieve higher test accuracy than training from scratch.
We focus on training GConv variants of ResNet-164 on CIFAR-100 by the same training from scratch schedule as~\cite{pytorch-classification}.
These variants use uniform $G$ numbers 2 and 4, and they may use channel shuffle~\cite{Zhang2017a} to exchange information among groups or not.
We also provide models with the same $G$ number produced by our pruning method.

Table~\ref{table:ablation:scratch} shows that overall training from scratch with channel shuffle performs better than other methods.
Comparing pruning with channel shuffle, we notice that the difference in train error is much larger than test error, which implies that the worse performance from the pruning method may due to its improper training setup, e.g., the number of training epochs is too limited.
But still, pruned models perform much better than their counterparts that are trained from scratch without channel permutation.
It shows that the permutation of channels after GConv is indeed an important architectural choice.


%


\subsection{Discussion}
We show that our method can balance the trade-off between accuracy and workload size induced by group configuration through our efficient pruning algorithm, and improve the trade-off by rapid exploration of configurations under given constraints.
Large and small models, small and large datasets are all covered.
Compared with the state-of-the-art structured pruning methods~\cite{Liu2017, Huang2017a, Peng2018}, our approach is significantly better regarding its exploration ability and efficiency.
Regarding model accuracy, we already perform on par with listed prior works, and we will try to surpass their results by tuning hyperparameters harder.

\section{Conclusion}

This paper proposes a novel pruning method that prunes a trained CNN model into one that is built on GConv.
We formulate the layer-wise pruning problem as finding optimal permutations to incorporate the structural constraints imposed by GConv, and we efficiently solve it through our heuristic algorithm.
We can further explore the best sparsity configuration of a whole model under specific inference budget constraints.
Empirical results show that with given sparsity, our pruning algorithm can achieve competitive accuracy as other prior work with a shorter pruning period; and the sparsity configuration exploration, which used to be intractable, can be efficiently performed by our method.

Future work includes exploring different importance criteria to improve the quality of explored models, tuning the pruning hyperparameters to achieve higher model accuracy after the fine-tuning phase, and investigating the contribution of the regularisation effect on the better accuracy from CondenseNet.
It is also possible to utilise NAS to improve our group configuration optimisation solution, since each group configuration basically determines an architecture.

\newpage

\section*{Acknowledgements}
The support of the United Kingdom EPSRC (grant numbers EP/P010040/1, EP/L016796/1, and EP/N031768/1),  Corerain, Intel, Maxeler and Xilinx is gratefully acknowledged.
We also thank James T. Wilson and Nadeen Gebara for their valuable suggestions.

{\small
\bibliographystyle{ieee}
\bibliography{references,custom}

\begin{thebibliography}{10}\itemsep=-1pt

\bibitem{Chen2019}
Y.~Chen, T.~Yang, X.~Zhang, G.~Meng, C.~Pan, and J.~Sun.
\newblock {DetNAS: Neural Architecture Search on Object Detection}.
\newblock pages 1--11, 2019.

\bibitem{Dong2017}
X.~Dong, S.~Chen, and S.~J. Pan.
\newblock {Learning to Prune Deep Neural Networks via Layer-wise Optimal Brain
  Surgeon}.
\newblock In {\em NeurIPS}, 2017.

\bibitem{Figurnov2015}
M.~Figurnov, A.~Ibraimova, D.~Vetrov, and P.~Kohli.
\newblock {PerforatedCNNs: Acceleration through Elimination of Redundant
  Convolutions}.
\newblock In {\em NeurIPS}, 2015.

\bibitem{Garfinkel1978}
R.~S. Garfinkel and K.~C. Gilbert.
\newblock {The Bottleneck Traveling Salesman Problem: Algorithms and
  Probabilistic Analysis}.
\newblock {\em Journal of the ACM}, 25(3):435--448, 1978.

\bibitem{Han2016}
S.~Han, H.~Mao, and W.~J. Dally.
\newblock {Deep Compression - Compressing Deep Neural Networks with Pruning,
  Trained Quantization and Huffman Coding}.
\newblock In {\em ICLR}, 2016.

\bibitem{Han2015}
S.~Han, J.~Pool, J.~Tran, and W.~J. Dally.
\newblock {Learning Both Weights and Connections for Efficient Neural
  Networks}.
\newblock In {\em NeurIPS}, 2015.

\bibitem{Hassibi1993a}
B.~Hassibi, D.~G. Stork, and G.~J. Wolff.
\newblock {Optimal brain surgeon and general network pruning}.
\newblock In {\em IEEE International Conference on Neural Networks}, 1993.

\bibitem{He2015}
K.~He, X.~Zhang, S.~Ren, and J.~Sun.
\newblock {Deep Residual Learning for Image Recognition}.
\newblock In {\em CVPR}, 2016.

\bibitem{He2016}
K.~He, X.~Zhang, S.~Ren, and J.~Sun.
\newblock {Identity Mappings in Deep Residual Networks}.
\newblock In {\em ECCV}, 2016.

\bibitem{Howard2017}
A.~G. Howard, M.~Zhu, B.~Chen, D.~Kalenichenko, W.~Wang, T.~Weyand,
  M.~Andreetto, and H.~Adam.
\newblock {MobileNets: Efficient Convolutional Neural Networks for Mobile
  Vision Applications}.
\newblock {\em CoRR}, 2017.

\bibitem{Hu2016}
H.~Hu, R.~Peng, Y.-W. Tai, and C.-K. Tang.
\newblock {Network Trimming: A Data-Driven Neuron Pruning Approach towards
  Efficient Deep Architectures}.
\newblock 2016.

\bibitem{Huang2017a}
G.~Huang, S.~Liu, L.~van~der Maaten, and K.~Q. Weinberger.
\newblock {CondenseNet: An Efficient DenseNet using Learned Group
  Convolutions}.
\newblock {\em CoRR}, abs/1711.0, 2017.

\bibitem{Huang2016a}
G.~Huang, Z.~Liu, L.~van~der Maaten, and K.~Q. Weinberger.
\newblock {Densely Connected Convolutional Networks}.
\newblock {\em CoRR}, abs/1608.0, 2016.

\bibitem{Huang2016}
J.~Huang, V.~Rathod, C.~Sun, M.~Zhu, A.~Korattikara, A.~Fathi, I.~Fischer,
  Z.~Wojna, Y.~Song, S.~Guadarrama, and K.~Murphy.
\newblock {Speed/accuracy trade-offs for modern convolutional object
  detectors}.
\newblock 2016.

\bibitem{Ioannou2016}
Y.~Ioannou, D.~Robertson, R.~Cipolla, and A.~Criminisi.
\newblock {Deep Roots: Improving CNN Efficiency with Hierarchical Filter
  Groups}.
\newblock In {\em CVPR}, 2017.

\bibitem{Krizhevsky2009}
A.~Krizhevsky.
\newblock {Learning Multiple Layers of Features from Tiny Images}.
\newblock Technical report, 2009.

\bibitem{Krizhevsky2012}
A.~Krizhevsky, I.~Sutskever, and G.~E. Hinton.
\newblock {ImageNet Classification with Deep Convolutional Neural Networks}.
\newblock In {\em NeurIPS}, 2012.

\bibitem{Lebedev2016}
V.~Lebedev and V.~Lempitsky.
\newblock {Fast ConvNets Using Group-Wise Brain Damage}.
\newblock In {\em CVPR}, pages 2554--2564, 2016.

\bibitem{LeCun1990}
Y.~LeCun, J.~S. Denker, and S.~a. Solla.
\newblock {Optimal Brain Damage}.
\newblock In {\em NeurIPS}, pages 598--605, 1990.

\bibitem{Li2017a}
H.~Li, A.~Kadav, I.~Durdanovic, H.~Samet, and H.~P. Graf.
\newblock {Pruning Filters for Efficient Convnets}.
\newblock In {\em ICLR}, 2017.

\bibitem{Liu2018a}
C.~Liu, B.~Zoph, M.~Neumann, J.~Shlens, W.~Hua, L.~J. Li, L.~Fei-Fei,
  A.~Yuille, J.~Huang, and K.~Murphy.
\newblock {Progressive Neural Architecture Search}.
\newblock {\em Lecture Notes in Computer Science (including subseries Lecture
  Notes in Artificial Intelligence and Lecture Notes in Bioinformatics)}, 11205
  LNCS:19--35, 2018.

\bibitem{Liu2017a}
H.~Liu, K.~Simonyan, O.~Vinyals, C.~Fernando, and K.~Kavukcuoglu.
\newblock {Hierarchical Representations for Efficient Architecture Search}.
\newblock In {\em ICLR}, 2018.

\bibitem{Liu2018c}
H.~Liu, K.~Simonyan, and Y.~Yang.
\newblock {DARTS: Differentiable Architecture Search}.
\newblock In {\em ICLR}, 2019.

\bibitem{Liu2017}
Z.~Liu, J.~Li, Z.~Shen, G.~Huang, S.~Yan, and C.~Zhang.
\newblock {Learning Efficient Convolutional Networks through Network Slimming}.
\newblock In {\em ICCV}, pages 2755--2763, 2017.

\bibitem{Liu2019}
Z.~Liu, M.~Sun, T.~Zhou, G.~Huang, and T.~Darrell.
\newblock {Rethinking the Value of Network Pruning}.
\newblock In {\em ICLR}, 2019.

\bibitem{Luo2017}
J.~H. Luo, J.~Wu, and W.~Lin.
\newblock {ThiNet: A Filter Level Pruning Method for Deep Neural Network
  Compression}.
\newblock In {\em ICCV}, 2017.

\bibitem{Mao2017}
H.~Mao, S.~Han, J.~Pool, W.~Li, X.~Liu, Y.~Wang, and W.~J. Dally.
\newblock {Exploring the Regularity of Sparse Structure in Convolutional Neural
  Networks}.
\newblock In {\em NeurIPS}, 2017.

\bibitem{Masana2017}
M.~Masana, J.~V.~D. Weijer, L.~Herranz, A.~D. Bagdanov, and J.~M. Alvarez.
\newblock {Domain-Adaptive Deep Network Compression}.
\newblock In {\em ICCV}, pages 4299--4307, 2017.

\bibitem{Molchanov2017}
P.~Molchanov, S.~Tyree, T.~Karras, T.~Aila, and J.~Kautz.
\newblock {Pruning Convolutional Neural Networks for Resource Efficient
  Transfer Learning}.
\newblock In {\em ICLR}, 2017.

\bibitem{Pan2016}
W.~Pan, H.~Dong, and Y.~Guo.
\newblock {DropNeuron: Simplifying the Structure of Deep Neural Networks}.
\newblock In {\em NeurIPS}, 2016.

\bibitem{Paszke2017}
A.~Paszke, G.~Chanan, Z.~Lin, S.~Gross, E.~Yang, L.~Antiga, and Z.~Devito.
\newblock {Automatic differentiation in PyTorch}.
\newblock In {\em NeurIPS}, 2017.

\bibitem{Peng2018}
B.~Peng, W.~Tan, Z.~Li, S.~Zhang, D.~Xie, and S.~Pu.
\newblock {Extreme Network Compression via Filter Group Approximation}.
\newblock In {\em ECCV}, 2018.

\bibitem{Real2018}
E.~Real, A.~Aggarwal, Y.~Huang, and Q.~V. Le.
\newblock {Regularized Evolution for Image Classifier Architecture Search}.
\newblock pages 52--54, 2018.

\bibitem{Russakovsky2015}
O.~Russakovsky, J.~Deng, H.~Su, J.~Krause, S.~Satheesh, S.~Ma, Z.~Huang,
  A.~Karpathy, A.~Khosla, M.~Bernstein, A.~C. Berg, and L.~Fei-Fei.
\newblock {ImageNet Large Scale Visual Recognition Challenge}.
\newblock {\em International Journal of Computer Vision}, 115(3):211--252,
  2015.

\bibitem{Russell}
S.~J. Russell and P.~Norvig.
\newblock {\em {Artificial Intelligence: A Modern Approach}}.

\bibitem{Sandler2018}
M.~Sandler, A.~Howard, M.~Zhu, A.~Zhmoginov, and L.-C. Chen.
\newblock {Inverted Residuals and Linear Bottlenecks: Mobile Networks for
  Classification, Detection and Segmentation}.
\newblock {\em CoRR}, 2018.

\bibitem{Tan2018}
M.~Tan, B.~Chen, R.~Pang, V.~Vasudevan, M.~Sandler, A.~Howard, and Q.~V. Le.
\newblock {MnasNet: Platform-Aware Neural Architecture Search for Mobile}.
\newblock 2018.

\bibitem{Tan2019}
M.~Tan and Q.~V. Le.
\newblock {EfficientNet: Rethinking Model Scaling for Convolutional Neural
  Networks}.
\newblock In {\em ICML}, 2019.

\bibitem{Wang2019a}
X.~Wang, M.~Kan, S.~Shan, and X.~Chen.
\newblock {Fully Learnable Group Convolution for Acceleration of Deep Neural
  Networks}.
\newblock In {\em CVPR}, pages 9049--9058, 2019.

\bibitem{Wen2016}
W.~Wen, C.~Wu, Y.~Wang, Y.~Chen, and H.~Li.
\newblock {Learning Structured Sparsity in Deep Neural Networks}.
\newblock In {\em NeurIPS}, 2016.

\bibitem{Williams1992}
R.~J. Williams.
\newblock {Simple statistical gradient-following algorithms for connectionist
  reinforcement learning}.
\newblock {\em Machine Learning}, 8(3-4):229--256, 1992.

\bibitem{Xie2018}
G.~Xie, J.~Wang, T.~Zhang, J.~Lai, R.~Hong, and G.-J. Qi.
\newblock {IGCV{\$}2{\$}: Interleaved Structured Sparse Convolutional Neural
  Networks}.
\newblock 2018.

\bibitem{Xie2018a}
G.~Xie, J.~Wang, T.~Zhang, J.~Lai, R.~Hong, and G.-J. Qi.
\newblock {Interleaved Structured Sparse Convolutional Neural Networks}.
\newblock In {\em CVPR}, 2018.

\bibitem{Xie2016}
S.~Xie, R.~Girshick, P.~Doll{\'{a}}r, Z.~Tu, and K.~He.
\newblock {Aggregated Residual Transformations for Deep Neural Networks}.
\newblock In {\em CVPR}, 2017.

\bibitem{pytorch-classification}
W.~Yang.
\newblock Pytorch classification.
\newblock \url{https://github.com/bearpaw/pytorch-classification}.

\bibitem{Yu2018}
R.~Yu, A.~Li, C.-F. Chen, J.-H. Lai, V.~I. Morariu, X.~Han, M.~Gao, C.-Y. Lin,
  and L.~S. Davis.
\newblock {NISP: Pruning Networks using Neuron Importance Score Propagation}.
\newblock In {\em CVPR}, 2018.

\bibitem{Yuan2006}
M.~Yuan and Y.~Lin.
\newblock {Model selection and estimation in regression with grouped
  variables}.
\newblock {\em Journal of the Royal Statistical Society. Series B: Statistical
  Methodology}, 68(1):49--67, 2006.

\bibitem{Zhang2017d}
T.~Zhang, G.-J. Qi, B.~Xiao, and J.~Wang.
\newblock {Interleaved Group Convolutions for Deep Neural Networks}.
\newblock In {\em ICCV}, 2017.

\bibitem{Zhang2017a}
X.~Zhang, X.~Zhou, M.~Lin, and J.~Sun.
\newblock {ShuffleNet: An Extremely Efficient Convolutional Neural Network for
  Mobile Devices}.
\newblock {\em CoRR}, 2017.

\bibitem{Zhao2018a}
R.~Zhao, Y.~Hu, J.~Dotzel, C.~De~Sa, and Z.~Zhang.
\newblock {Building Efficient Deep Neural Networks with Unitary Group
  Convolutions}.
\newblock 2018.

\bibitem{Zoph2016}
B.~Zoph and Q.~V. Le.
\newblock {Neural Architecture Search with Reinforcement Learning}.
\newblock {\em CoRR}, pages 1--16, 2016.

\bibitem{Zoph2017}
B.~Zoph, V.~Vasudevan, J.~Shlens, and Q.~V. Le.
\newblock {Learning Transferable Architectures for Scalable Image Recognition}.
\newblock 2017.

\end{thebibliography}


\begin{thebibliography}{10}\itemsep=-1pt

\bibitem{condensenet}
Condensenet repository.
\newblock \url{https://github.com/ShichenLiu/CondenseNet}.

\bibitem{pytorch-mobilenet}
Pytorch mobilenet.
\newblock \url{https://github.com/marvis/pytorch-mobilenet}.

\bibitem{torchvision}
Torchvision models.
\newblock \url{https://pytorch.org/docs/master/torchvision/models.html}.

\bibitem{He2015}
K.~He, X.~Zhang, S.~Ren, and J.~Sun.
\newblock {Deep Residual Learning for Image Recognition}.
\newblock In {\em CVPR}, 2016.

\bibitem{He2016}
K.~He, X.~Zhang, S.~Ren, and J.~Sun.
\newblock {Identity Mappings in Deep Residual Networks}.
\newblock In {\em ECCV}, 2016.

\bibitem{Howard2017}
A.~G. Howard, M.~Zhu, B.~Chen, D.~Kalenichenko, W.~Wang, T.~Weyand,
  M.~Andreetto, and H.~Adam.
\newblock {MobileNets: Efficient Convolutional Neural Networks for Mobile
  Vision Applications}.
\newblock {\em CoRR}, 2017.

\bibitem{Huang2017}
G.~Huang, D.~Chen, T.~Li, F.~Wu, L.~van~der Maaten, and K.~Q. Weinberger.
\newblock {Multi-Scale Dense Networks for Resource Efficient Image
  Classification}.
\newblock In {\em ICLR}, 2018.

\bibitem{Huang2017a}
G.~Huang, S.~Liu, L.~van~der Maaten, and K.~Q. Weinberger.
\newblock {CondenseNet: An Efficient DenseNet using Learned Group
  Convolutions}.
\newblock {\em CoRR}, abs/1711.0, 2017.

\bibitem{Liu2017}
Z.~Liu, J.~Li, Z.~Shen, G.~Huang, S.~Yan, and C.~Zhang.
\newblock {Learning Efficient Convolutional Networks through Network Slimming}.
\newblock In {\em ICCV}, pages 2755--2763, 2017.

\bibitem{Luo2017}
J.~H. Luo, J.~Wu, and W.~Lin.
\newblock {ThiNet: A Filter Level Pruning Method for Deep Neural Network
  Compression}.
\newblock In {\em ICCV}, 2017.

\bibitem{Molchanov2017}
P.~Molchanov, S.~Tyree, T.~Karras, T.~Aila, and J.~Kautz.
\newblock {Pruning Convolutional Neural Networks for Resource Efficient
  Transfer Learning}.
\newblock In {\em ICLR}, 2017.

\bibitem{Paszke2017}
A.~Paszke, G.~Chanan, Z.~Lin, S.~Gross, E.~Yang, L.~Antiga, and Z.~Devito.
\newblock {Automatic differentiation in PyTorch}.
\newblock In {\em NeurIPS}, 2017.

\bibitem{Peng2018}
B.~Peng, W.~Tan, Z.~Li, S.~Zhang, D.~Xie, and S.~Pu.
\newblock {Extreme Network Compression via Filter Group Approximation}.
\newblock In {\em ECCV}, 2018.

\bibitem{Wang2016b}
Y.~Wang, V.~I. Morariu, and L.~S. Davis.
\newblock {Learning a Discriminative Filter Bank within a CNN for Fine-grained
  Recognition}.
\newblock In {\em CVPR}, pages 4148--4157, 2018.

\bibitem{pytorch-classification}
W.~Yang.
\newblock Pytorch classification.
\newblock \url{https://github.com/bearpaw/pytorch-classification}.

\bibitem{Yu2018}
R.~Yu, A.~Li, C.-F. Chen, J.-H. Lai, V.~I. Morariu, X.~Han, M.~Gao, C.-Y. Lin,
  and L.~S. Davis.
\newblock {NISP: Pruning Networks using Neuron Importance Score Propagation}.
\newblock In {\em CVPR}, 2018.

\bibitem{Zhao2018a}
R.~Zhao, Y.~Hu, J.~Dotzel, C.~{De Sa}, and Z.~Zhang.
\newblock {Building Efficient Deep Neural Networks with Unitary Group
  Convolutions}.
\newblock 2018.

\bibitem{Zhu2017}
M.~Zhu and S.~Gupta.
\newblock {To prune, or not to prune: exploring the efficacy of pruning for
  model compression}.
\newblock {\em CoRR}, 2017.

\end{thebibliography}
}

\end{document}


\title{Efficient Structured Pruning Method for Group Convolution\\
 (Supplementary Material)}

\author{First Author\\
Institution1\\
Institution1 address\\
{\tt\small firstauthor@i1.org}
\and
Second Author\\
Institution2\\
First line of institution2 address\\
{\tt\small secondauthor@i2.org}
}

\maketitle

\tableofcontents
\newpage

\section{Explanation of the Heuristic Algorithm for Layer-wise Pruning}

This heuristic algorithm is designed to solve the permutation based pruning problem.
We use this algorithm to move important kernels to the diagonal blocks so that pruning those outside the diagonal will hurt less to the test accuracy.
In this section, we intend to clarify some points that may not be thoroughly explained in the paper.

\paragraph{Notations and terminologies.}
Table~\ref{table:notation} summarises all the notations used to describe the heuristic algorithm
and \eqref{eq:notation} contains a list of equations that relate different notations.

\begin{table}[h]
  \centering
  \caption{Notations used to describe the heuristic algorithm. The shape of each tensor related to each notation is listed as }
  \label{table:notation}
  \begin{tabular}{ll} 
    \hline
    {\bf Notation} & {\bf Description} \\\hline
    $\X$           & Input feature map $(\CI, \HH, \WW)$ \\
    $\Y$           & Output feature map $(\CO, \HH, \WW)$ \\
    $\W$           & Pre-trained weights $(\CO, \CI, \KH, \KW)$ \\
    \hline
    $\II$          & Permutation of input channel indices $(\CI)$ \\
    $\IO$          & Permutation of output channel indices $(\CO)$ \\
    \hline
    $\tilde{\X}$   & Permuted input feature map $(\CI, \HH, \WW)$ \\
    $\tilde{\Y}$   & Permuted output feature map $(\CO, \HH, \WW)$ \\
    $\tilde{\W}$   & Permuted weights $(\CO, \CI, \KH, \KW)$ \\
    \hline
    $\XS$          & Partitioned input feature map $(\GG, \CI/\GG, \HH, \WW)$ \\
    $\YS$          & Partitioned output feature map $(\GG, \CO/\GG, \HH, \WW)$ \\
    $\WS$          & Partitioned weights $(\GG, \GG, \CO/\GG, \CI/\GG, \KH, \KW)$ \\
    $\tilde{\WS}$  &
       Permuted and then partitioned weights $(\GG, \GG, \CO/\GG, \CI/\GG, \KH, \KW)$ \\
    \hline
    $\X^g$         &
      A partition of input feature map belongs to group $g$ $(\CI/\GG, \HH, \WW)$ \\
    $\Y^g$         & 
      A partition of output feature map belongs to group $g$ $(\CO/\GG, \HH, \WW)$ \\
    $\W^g$         & 
      A partition of weights belongs to input group $g$ and output group $g$
      $(\CO/\GG, \CI/\GG, \KH, \KW)$ \\
    $\W^{g_f,g_c}$         & 
      A partition of weights belongs to input group $g_c$ and output group $g_f$
      $(\CO/\GG, \CI/\GG, \KH, \KW)$ \\
    \hline
  \end{tabular}
\end{table}

\begin{equation}\label{eq:notation}
  \begin{aligned}
    \Y_f &= \sum_{c=1}^{\CI} \X_c * \W_{fc} \quad \forall\ 1\leq f\leq \CO
    &\quad \text{Standard convolution definition} \\
    \tilde{\X}_c &= \X_{\II(c)}  \quad \forall\ 1\leq c\leq \CI
    & \quad \text{Permute channels of the input feature map} \\
    \Y_f &= \tilde{\Y}_{\IO(f)}  \quad \forall\ 1\leq f\leq \CO
    & \quad \text{Permute channels of the output feature map} \\
    \tilde{\W}_{fc} &= \W_{\IO(f)\II(c)}
    \quad \forall\ 1\leq f\leq \CO
    \quad \forall\ 1\leq c\leq \CI
    & \quad \text{Permute weights by input and output channels} \\
    \XS_{g,c} &= \X_{\frac{(g-1)\CI}{\GG}+c}
    &\quad \text{Partition input feature map along channels} \\
    \YS_{g,f} &= \Y_{\frac{(g-1)\CO}{\GG}+f}
    &\quad \text{Partition output feature map along channels} \\
    \WS_{g_f,g_c,f,c} &= \W_{\frac{(g_f-1)\CO}{\GG}+f, \frac{(g_c-1)\CI}{\GG}+c}
    &\quad \text{Partition weights channels by group} \\
    \W^g &= \WS_{g,g} \\
    \W^{g_f,g_c} &= \WS_{g_f,g_c} \\
    \X^g &= \XS_{g} \\
    \Y^g &= \YS_{g} \\
  \end{aligned}
\end{equation}

\paragraph{Pseudo-code.}
In the paper we describe the heuristic algorithm for layer-wise pruning in a figure form accompanied by text description.
This section illustrates the steps of the algorithm more formally as pseudo-code (Algorithm~\ref{alg:greedy}).

In this algorithm, we first initialise the permutation indices in their original sequences, i.e., from 1 to the number of channels.
We also compute the L2 norm of all kernels and assign them to a matrix $\hat{\W}$.
The core loop iterates a index of group $g$ from $G$ to $1$.
In each iteration, we compute the range of channels that can be sorted, denoted as $\bar{f}$ and $\bar{c}$, which exclude those that are covered by those blocks that have been frozen.
After that, within the given ranges, we compute the keys for sorting $\SKf$ and $\SKc$ and calls the $Sort$ function (lines \ref{alg:greedy:sort_in} and \ref{alg:greedy:sort_out}), which is an analogy of NumPy \texttt{argsort}\footnote{\url{https://docs.scipy.org/doc/numpy-1.15.1/reference/generated/numpy.argsort.html}}:
it takes an array, a function to calculate the key for sorting at each position, and a range of sorting, and returns the indices of the sorted array.
The returned values from $Sort$, which are indices of channels in their sorted order, will be used to assign $\II$ and $\IO$.

\begin{algorithm}[ht]
  \small
  \caption{
    The layer-wise heuristic pruning algorithm. The $Sort(\I,\sigma, range)$ function sorts indices $\I$ in an increasing order defined by a key $\sigma$ within a given range (denoted as $range$ which is a set of indices), and returns indices of the sorted array.}
  \label{alg:greedy}

  \DontPrintSemicolon

  \SetKwInOut{Input}{Input}
  \SetKwInOut{Output}{Output}
  \SetKwData{cols}{C}
  \SetKwData{rows}{R}
  \SetKwFunction{sort}{sort}

  \KwData{Weights $\W\in\mathbb{R}^{\CO\times\CI\times K_h\times K_w}$, a given number of groups $\GG$.}
  \KwResult{A pair of permutation indices $\IO$ and $\II$}

  \Begin{
    \tcp{Initialise channel permutation indices in their original order.}
    $\II, \IO \gets \{1,\dots,\CI\}, \{1,\dots,\CO\}$\;

    \tcp{Initialise each entry with the L2 norm of each kernel.}
    \For{$f\in\{1,\dots, \CO \}$, $c\in\{1,\dots, \CI \}$}{
      $\widehat{\W}_{fc} \gets \norm{\W_{fc}}$\;
    }

    \tcp{Iterate every diagonal block.}
    \For{$g \gets \{G, \dots, 2, 1\}$}{\label{alg:greedy:block_loop}
      \tcp{Range of channels covered in $\SKf$ and $\SKc$}
      $\bar{f} \gets \left\{
        \frac{(g-1)}{G}\CO + 1,\dots,\frac{g}{G}\CO\right\}$\;
      $\bar{c} \gets \left\{
        \frac{(g-1)}{G}\CI + 1,\dots,\frac{g}{G}\CI\right\}$\;

      \tcp{Compute $\SKc$ and sort $\II$.}
      \For{$1\leq c \leq \frac{g}{G}\CI$}{
        $\SKc(c, \GG) \gets
         \sum\limits^{f\in\bar{f}} \widehat{\W}_{\IO(f)\II(c)}$
      }
      $\II \gets Sort(\II, \SKc, \bar{c})$\;
      \label{alg:greedy:sort_in}

      \tcp{Compute $\SKf$ and sort $\IO$.}
      \For{$1\leq f \leq \frac{g}{G}\CO$}{
        $\SKf(f, \GG) \gets
         \sum\limits^{c\in\bar{c}} \widehat{\W}_{\IO(f)\II(c)}$
      }
      $\IO \gets Sort(\IO, \SKf, \bar{f})$\;
      \label{alg:greedy:sort_out}
    }
  }
\end{algorithm}

\begin{figure}[ht]
  \centering
  \begin{subfigure}[b]{0.4\textwidth}
    \centering
    \includegraphics[width=\textwidth]{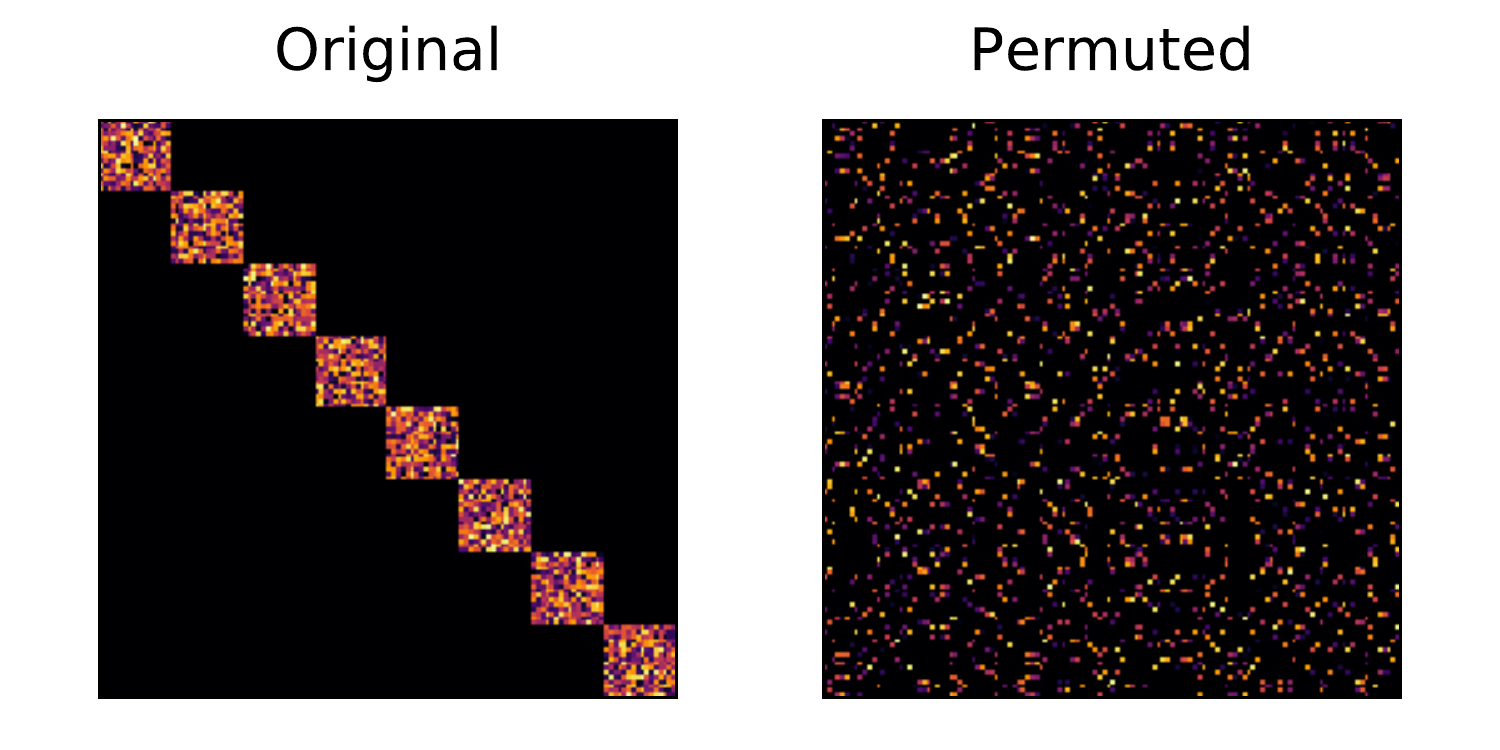}
  \end{subfigure}
  ~
  \begin{subfigure}[b]{0.4\textwidth}
    \centering
    \includegraphics[width=\textwidth]{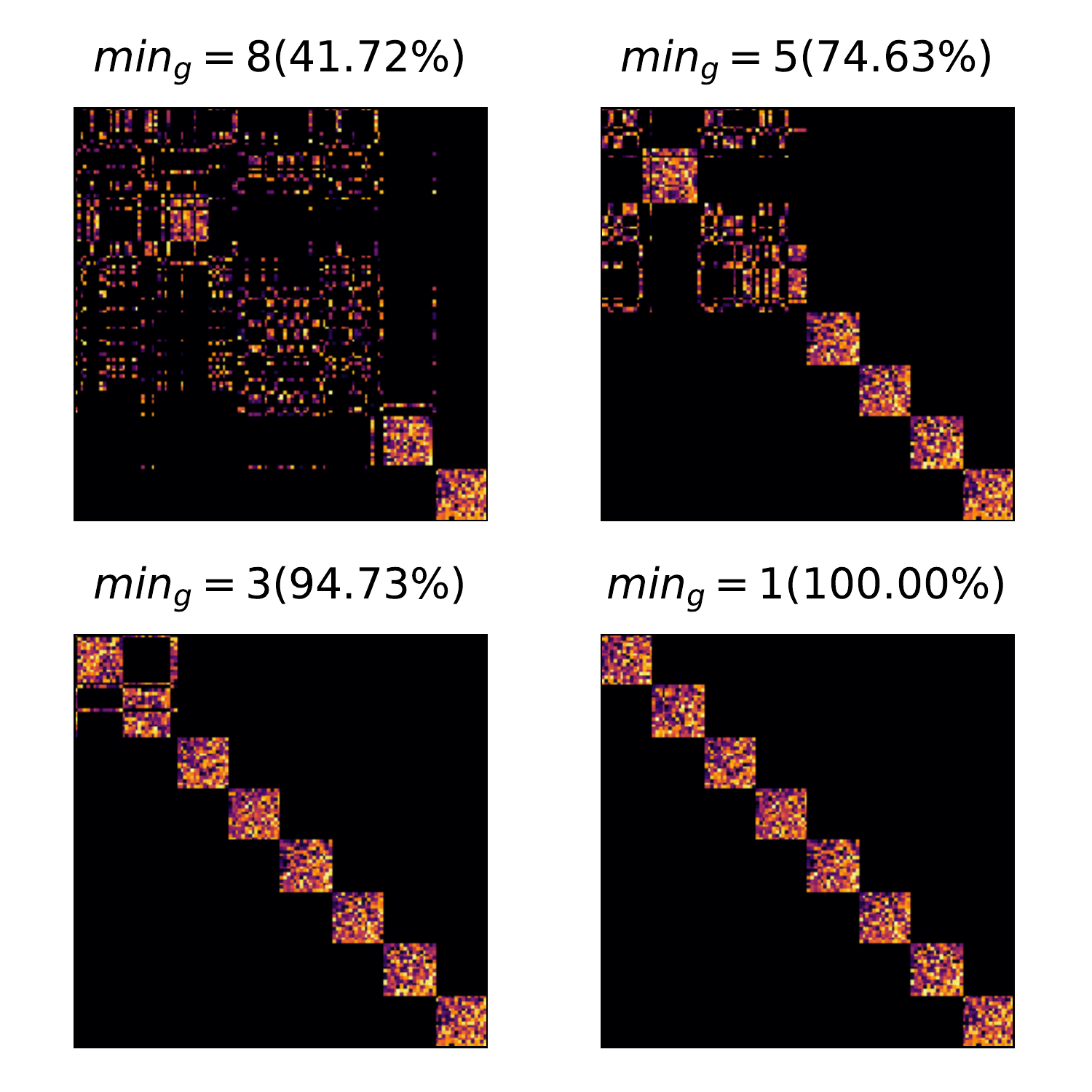}
  \end{subfigure}
  \caption{Illustration of several steps during running the algorithm. $min_g$ denotes the lower bound of $g$, which is the point that we end the algorithm and extract the output.}
  \label{fig:step_by_step}
\end{figure}

\paragraph{Illustration.}
\figref{fig:step_by_step} illustrates the procedure of running this algorithm step by step.
The input to this algorithm in this example is originally a block-diagonal matrix and permuted randomly.
Then we run the heuristic algorithm on the permuted matrix.
In this illustration, at the first step, we can recover the bottom-right block.
After that, we continuously recover diagonal blocks one by one, until we reach the top-left block.
This example is also one of the samples being used in the random testing mentioned in the paper.

\begin{figure}[ht]
  \centering
  \includegraphics[width=0.6\textwidth]{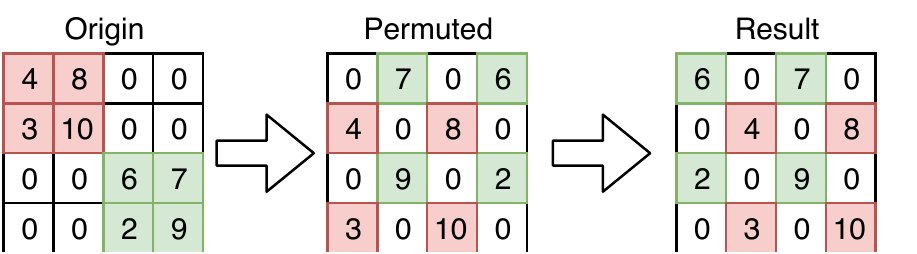}
  \caption{The worst case scenario. The origin block-diagonal matrix is the leftmost one, we permute it into the one in the middle, and running our algorithm results in the rightmost one.
  We colour non-zero entries.}
  \label{fig:worst_case}
\end{figure}

\paragraph{Worst-case senario.}
Ideally, this algorithm can work smoothly as the example mentioned above (\figref{fig:step_by_step}), which is also justified by the statistical result collected from 10k random test cases.
But there are some cases that this algorithm will fail to handle, especially when knowing the global information is critical to achieve the optimal result.
One typical worst-case scenario is illustrated in \figref{fig:worst_case}.
In this case, we have a $4\times 4$ weight matrix and $G$ is 2.
When running the sorting algorithm, based on the key $\SKc$, we will move the 2nd and 4th columns towards the right because 9 and 10 are the two largest values in the bottom two rows.
But the optimal solution is to move the 1st and 3rd columns to the right, so that we can keep non-zero elements compactly within blocks.
This case implies that $\SKc$ and $\SKf$ that consider only local information cannot be helpful to reach a good state regarding global information.
\section{Explanation of the Local Search Algorithm}

We present the local search algorithm in the paper, which is designed to find an optimal sparsity configuration across all convolution layers.
This section presents more details on this topic.

\paragraph{Notations and terminologies.}
Suppose the given model has $L$ convolution layers, for each of them, we can assign a value that specifies the sparsity, which is the number of groups in our case.
Let $\G$ denote the sparsity of all convolution layers, i.e., $\G_l$ is the number of groups for the $l$-th layer, $1\leq l\leq L$.
Our objective is to find the $\G$, by using which we can get a pruned model that has the maximal test accuracy.

The search space of $\G$ grows exponentially while $L$ increases, therefore, we propose the local search algorithm to find a near-optimal sparsity configuration.
First of all, we notice that for each layer, the candidates for $\G_l$ are the common factors of $\CI$ and $\CO$, denoted by $\bar{\G}^l$.
Candidates are sorted in an decreasing order in $\bar{\G}^{(l)}$ so that $\bar{\G}^{(l)}_i$ is the $i$-th large candidate.
Since we cannot directly predict the test accuracy of a sparse configuration, we reuse the L2 norm criterion to estimate the cost of pruning.

\paragraph{Pseudo-code.}
The pseudo-code for this local search algorithm is listed in Algorithm~\ref{alg:local_search}.
Instead of using $\G$ to represent the state, we use $\ST$ which indicates the indices of candidates from $\bar{\G}$ that we should use to assign to $\G$. 
For example, we can deduce $\G_l$ by $\bar{\G}^{(l)}_{\ST_l}$.
All elements of $\ST$ are initialised as 1 so that all layers start from their largest possible numbers of groups, which has the highest cost and the lowest inference workload.
In the main loop of this algorithm, we try to find a layer that can achieve the minimal cost by selecting the nearest smaller candidate ($\ST_l + 1$ in line~\ref{alg:local_search:state}).
We also exclude those layers that have reached the largest $\ST_l$, which is the cardinality of its candidate set $|\bar{\G}^{(l)}|$.
The loop terminates when the total numbers of parameters and operations exceed the boundary defined by $\Nparmax$ and $\Nopsmax$.
The numbers of parameters and operations are computed by \texttt{NumOfParams} and \texttt{NumOfOps} respectively.
Note that \texttt{NumOfOps} needs to take $\YS^{(l)}$ to get the size of the output to compute the total number of operations.

\begin{algorithm}[ht]
  \caption{The local search algorithm.}
  \label{alg:local_search}

  \DontPrintSemicolon
  \SetKwRepeat{Do}{do}{while}
  \SetKwFunction{NumOfParams}{NumOfParams}
  \SetKwFunction{NumOfOps}{NumOfOps}

  $\ST\gets \{ 1 \}_{l=1}^L$\;

  \Do{$\Npar\leq\Nparmax \wedge \Nops\leq\Nopsmax$}{
    $l^* \gets \argmin\limits_{l}\left\{\,
      \mathsf{cost}\left(
        \WS^{(l)}, \bar{\G}^{(l)}_{\ST_l + 1} \right)\,
      \middle|\,
      \forall\ 1\leq l\leq L \wedge \ST_l <  | \bar{\G}^{(l)} |\,
      \right\}$\label{alg:local_search:state}\;

    $\ST_{l^*}\gets\ST_{l^*} + 1$\;

    $\Npar\gets
     \sum_{l=1}^L \NumOfParams(\WS^{(l)}, \bar{\G}^{(l)}_{\ST_l})$\;
    $\Nops\gets
     \sum_{l=1}^L \NumOfOps(\YS^{(l)}, \WS^{(l)}, \bar{\G}^{(l)}_{\ST_l})$\;
  }

  \Return $\ST$\;
\end{algorithm}

\paragraph{Problems from the cost estimator.}
This local search algorithm, as well as the heuristic layer-wise pruning algorithm, are based on the magnitude based kernel importance criterion.
As demonstrated by some recent papers~\cite{Molchanov2017,Yu2018,Luo2017}, this criterion is not very informative on the actual contribution from kernels to the accuracy.
In our future work, we will look at different cost estimation method, especially the first-order gradient based one as in~\cite{Molchanov2017}.
It is worthwhile to mention that the overall framework of our algorithm will not change if we apply this method, only collecting the first-order gradients and replacing L2 norm of kernels with these new values should be performed.
\section{Group Convolution Implementation}

This section lists two code snippets that demonstrate how we implement GConv.
In our paper, we present two different approaches of implementing GConv, one is by directly mapping from the permutation based GConv definition, the other one uses sparse matrix multiplication.

The direct mapping implementation will be useful when continuously calling multiple convolution kernels does not incur much overhead, while this is not applicable to modern GPU platforms.
We list the PyTorch code that implements this version in Listing~\ref{lst:gconv}.

\begin{listing}[ht]
\caption{
  This direct GConv implementation is built as a PyTorch~\cite{Paszke2017} \texttt{Module}, which performs computation in the \texttt{forward} method.
  We permute (\texttt{index\_select}) the input \texttt{x} by $\II$ (\texttt{ind\_in}) along the channel axis and the output by $\IO$ (\texttt{ind\_out}).
  Convolution groups (\texttt{conv}) takes input groups (\texttt{split}) and concatenate (\texttt{cat}) their results\label{lst:gconv}.}

\begin{minted}[]{python}
class GroupConv2d(Module):
  def forward(self, x):
    dim = 1 # index of the channel axis
    gs = index_select(x, dim, self.ind_in).split(self.G, dim=dim)
    os = [self.conv[i](g) for i, g in enumerate(gs)]
    return index_select(cat(os), dim, self.ind_out)
\end{minted}
\end{listing}

To efficiently utilise modern massive parallel processors, e.g., GPUs, we propose another sparse matrix multiplication (SpMM) based implementation (Listing~\ref{lst:gconv:sparse}).
As far as we know, the SpMM operation provided by PyTorch uses COO format normally, and if the sparse tensor is coalesced, then it will apply the CSR format.
The CSR format is suitable for our case, since our sparse weights have identical number of non-zero kernels on each row (output channel).

\begin{listing}[ht]
\caption{When using this module, we cannot simply assign weight tensor to this module. We provide a \texttt{update\_weight} function that turns the weight to be assigned into sparse tensor. The computation is done by \texttt{torch.sparse.mm}, the sparse matrix multiplication API provided by PyTorch v1.0.1\label{lst:gconv:sparse}.}

\begin{minted}[]{python}
class SparseGroupConv2d(Module):
  def update_weight(self, W):
    i = torch.nonzero(W).t()
    v = W[i[0], i[1]]
    self.weight.data = torch.sparse.FloatTensor(
      i, v, torch.Size(W.shape)).coalesce()

  def forward(self, x):
    return torch.sparse.mm(self.weight, x) 
\end{minted}
\end{listing}

\section{Details of the Experiments}

This section demonstrates the experimental results with more details and in-depth analysis.

\subsection{Reproducibility}

First of all, we list most of the necessary information here for purposes of reproducing our results.
Due to the double-blind review procedure we cannot disclose our code repository, so that not all information can be listed here.
We will demonstrate the actual training instructions once our codebase is open sourced.

\begin{table}[ht]
  \centering
  \caption{Baseline models collected for CIFAR-10/100 and ImageNet.}
  \label{table:baseline}
  \begin{tabular}{l|c|l|cc}
    \hline
     & & & \multicolumn{2}{c}{Top-1 Error(\%)} \\\cline{4-5}
    Model & Source & Dataset & Reported & Achieved \\
    \hline
    
    \multirow{2}{*}{ResNet-110~\cite{He2015}} 
    & \multirow{2}{*}{\cite{pytorch-classification}}
      & CIFAR-10  & 6.43~\cite{He2015}                  & 6.15 \\ 
    & & CIFAR-100 & 28.86~\cite{pytorch-classification} & 26.66 \\
    \hline

    \multirow{2}{*}{ResNet-164~\cite{He2016}}
    & \multirow{2}{*}{\cite{pytorch-classification}}
      & CIFAR-10  & 5.46~\cite{He2016}  & 5.29  \\
    & & CIFAR-100 & 24.33~\cite{He2016} & 22.83 \\\hline

    \multirow{2}{*}{DenseNet-86~\cite{Huang2017}} 
    & \multirow{2}{*}{\cite{condensenet}}
      & CIFAR-10  & ---  & 4.44  \\
    & & CIFAR-100 & ---  & 20.57 \\
    \hline

    ResNet-18 \cite{He2015}
      & \multirow{4}{*}{\cite{torchvision}} & \multirow{4}{*}{ImageNet}      
                                & 30.24 & --- \\
    ResNet-34 \cite{He2015} & & & 26.70 & --- \\
    ResNet-50 \cite{He2015} & & & 23.85 & --- \\
    ResNet-101\cite{He2015} & & & 22.63 & --- \\
    \hline

    MobileNet-224~\cite{Howard2017}
      & \cite{pytorch-mobilenet} & ImageNet
      & 70.6~\cite{Howard2017} & 69.52~\cite{pytorch-mobilenet}\\
    \hline
  \end{tabular}
\end{table}

\subsubsection{Baseline Models}

Models input to our method are collected mainly from the internet, via downloading or training by following the instructions from publicly available repositories.
How baseline models are collected or trained for CIFAR-10/100 and ImageNet are covered in Table~\ref{table:baseline}:

\begin{enumerate}[itemsep=0em]
  \item ResNet-110/164 are trained from scratch based on the pytorch-classification repository~\cite{pytorch-classification} by following their own training schedule.
  We have achieved the accuracy reported by either the original papers~\cite{He2015,He2016} or the repository.
  \item DenseNet-86 is constructed from CondenseNet-86 by changing all GConv layers to normal convolution layers, and is trained by the same training schedule as CondenseNet-86 (300 epochs, learning rate starts from 0.1 and applies cosine decay) excluding the pruning step.
  There is no published result on this model so that we just list ours.
  \item ResNet-18/34/50/101 are all downloaded from the TorchVision model zoo~\cite{torchvision}.
  \item MobileNet-224 is downloaded from a open repository~\cite{pytorch-mobilenet}. Its accuracy is slightly lower than the official result~\cite{Howard2017}. 
\end{enumerate}

Baseline models (ResNet-50 and MobileNet-224) of CUB-200 are fine-tuned from corresponding ImageNet baseline models.
Due to the limitation of time, we can only arbitrarily select hyperparameters for this procedure:
for ResNet-50, we use 50 epochs and a fixed learning rate $1e^{-3}$;
and for MobileNet-224, we borrow the hyperparameters from~\cite{He2015} and fine-tune it for 90 epochs with a learning rate starting at 0.01 and decays by a factor of 0.1 for every 30 epochs.
Also, we keep the weights of convolution layers unchanged during this fine-tuning procedure to minimise the training effort in this phase.
However, this approach may decrease the accuracy of the pruning cost estimation since we still estimate by the same weights as those trained for ImageNet, not CUB-200.
It is possible that by finding a better set of hyperparameters we can have a better starting point for further pruning.
We will investigate this in the future.

\subsubsection{Pruned Models}

A pruned model is defined by its base topology and a sparsity configuration, e.g., a pruned ResNet-50 model can be specified by the original ResNet-50 topology and the number of groups for all of its layers.
The sparsity configuration is either assigned by an ad-hoc rule, e.g., using a uniform number of groups for all layers, or solved by the local search algorithm within the model pruner.
Once we have the sparsity configuration, we can turn all convolution layers into their corresponding GConv counterparts with numbers of groups as described in the configuration.

We construct each GConv layer in the following steps:

\begin{enumerate}[itemsep=0em]
  \item Collecting the number of groups from the sparsity configuration and weights from the pre-trained model.
  \item Running the layer-wise pruner to determine which weights should be pruned.
  \item Building a mask on weights that keeps the values of weights that are retained and outputs zero for those pruned.
  \item Fine-tuning this pruned model built on masks for $1/3$ of the original training workload.
  \item Export the fine-tuned model to a deployable one.
\end{enumerate}

Instead of using the direct implementation of GConv, in step 3 we choose to build a masked version of convolution layers that in fact implements GConv to reduce the fine-tuning period.
The forward and backward computation of a masked convolution layer is faster than any implementation of a GConv layer.
This technique has been used in many other prior papers~\cite{Zhu2017,Huang2017a,Liu2017}.

We run step 4 on this masked model.
The difference in hyperparameters between training baseline models and fine-tuning pruned models are summarised in Table~\ref{table:hyper}.
Most of them are constrained by the $1/3$ workload rule, and there are some exceptions:

\begin{enumerate}[itemsep=0em]
  \item We discard the original cosine decay and use a multi-step decay similar to ResNet-110/164.
  \item We still keep the original workload for CUB-200 mainly because it is a rather small dataset while the task is still difficult. Therefore, in this case we loose the epochs constraints.
\end{enumerate}

When exporting the models, we replace every masked convolution layer with the desired GConv implementation.
If the implementation that is directly mapped from the GConv definition is the target, we can figure out which weights should belong to which group based on the mask.
This is done by permuting channels to move all the remaining weights to diagonal blocks.
Then we can extract the weights for all groups, which are contents in diagonal blocks, and the permutation indices, which are used to move remaining weights to the diagonal.
This implementation is shown in Listing~\ref{lst:gconv}.
When dealing with the sparse matrix implementation, we just element-wise multiply the mask with the weights and create a sparse matrix out of the result.
Note that we only map $1\times 1$ GConv to sparse matrix multiplication, doing so for $3\times 3$ GConv is costly since DNN libraries are optimised deeply for this case.
In this case, $3\times 3$ GConv layers are kept as their standard convolution implementation.
More details can be found in our codebase to be released in the future.

\begin{table}[ht]
  \centering
  \small
  \caption{Hyperparameters for training baseline models and fine-tuning pruned models.}
  \label{table:hyper}
  \begin{tabular}{l|l|ccp{8em}|ccp{8em}}
    \hline
              & & \multicolumn{3}{c|}{Training baseline models}
                & \multicolumn{3}{c}{Fine-tuning pruned models} \\\cline{3-8}
    Dataset
    & Model Name  & Epochs & LR & LR decay 
                  & Epochs & LR & LR decay \\\hline
    \multirow{3}{*}{CIFAR}
    & ResNet-110  & 164    & 0.1      & Scale by 0.1 at 81 and 122 epochs
                  & 60     & $5e^{-3}$& Scale by 0.1 at 30 and 45 epochs \\\cline{2-8}
    & ResNet-164  & 164    & 0.1      & Scale by 0.1 at 81 and 122 epochs
                  & 60     & $5e^{-3}$& Scale by 0.1 at 30 and 45 epochs \\\cline{2-8}
    & DenseNet-86 & 300    & 0.1      & Cosine decay 
                  & 100    & $5e^{-3}$& Scale by 0.1 at 50 and 75 epochs \\\hline
    \multirow{4}{*}{ImageNet}
    & ResNet-18   & 90     & 0.1      & Scale by 0.1 every 30 epochs
                  & 30     & $1e^{-3}$& Scale by 0.1 every 10 epochs \\\cline{2-8}
    & ResNet-34   & 90     & 0.1      & Scale by 0.1 every 30 epochs
                  & 30     & $1e^{-3}$& Scale by 0.1 every 10 epochs \\\cline{2-8}
    & ResNet-50   & 90     & 0.1      & Scale by 0.1 every 30 epochs
                  & 30     & $1e^{-3}$& Scale by 0.1 every 10 epochs \\\cline{2-8}
    & ResNet-101  & 90     & 0.1      & Scale by 0.1 every 30 epochs
                  & 30     & $1e^{-3}$& Scale by 0.1 every 10 epochs \\\hline
    \multirow{2}{*}{CUB-200}
    & ResNet-50     & 50     & $1e^{-3}$& Fixed 
                    & 90     & 0.01     & Scale by 0.1 every 30 epochs \\\cline{2-8}
    & MobileNet-224 & 90     & 0.01     & Scale by 0.1 every 30 epochs
                    & 90     & 0.01     & Scale by 0.1 every 30 epochs \\\hline
  \end{tabular}
\end{table}

\subsubsection{Software and Hardware}

We mainly evaluate the inference speed on a rack server (denoted by Server) and NVIDIA Jetson TX2 (denoted by TX2).
The Server runs Python 3.6.3 (built by Anaconda\footnote{\url{https://www.anaconda.com/}}) and installs PyTorch from its maintained stable pre-builds\footnote{\url{https://pytorch.org/get-started/locally/}} of version 1.0.1.
This build is shipped with built-in CUDA (version 10.0) and cuDNN (7.4).
For TX2, we follow instructions from here\footnote{\url{https://github.com/andrewadare/jetson-tx2-pytorch}} to build PyTorch 1.0.1 locally.
All the dependencies are installed by the package manager on Ubuntu LTS 16.04.
On this platform, the version of CUDA is 9.0 and cuDNN is 7.0.

\subsection{Explanation of Results}

This section reviews the experiments listed in the paper and gives in-depth analysis.

\subsubsection{Training Budget} 

\begin{table}[ht]
  \centering
  \small
  \caption{
    Comparison of the training budget among pairs of models that have similar top-1 test error.}
  \label{table:train_budget}
  \begin{tabular}{l|ccccp{18em}}
    \hline
    Model & Dataset & Err. (\%) & \# params. & \# FLOPS. & Training Budget \\\hline
    ResNet-164 ($G=2$)       & CIFAR-100 & 23.66 & 0.91M & 257M 
                             & Pruning and fine-tuning (60 epochs) \\
    ResNet-164 (60\% pruned)~\cite{Liu2017}
                             & CIFAR-100 & 23.91 & 1.21M & 247M
                             & Regularisation (160 epochs), pruning, fine-tuning (160 epochs) \\
    \hline
    DenseNet-86 ($G=4$)      & CIFAR-100 & 25.96 & 0.59M & 132.7M
                             & Pruning and fine-tuning (100 epochs)\\ 
    DenseNet-86 ($G=4$)~\cite{Huang2017a}
                             & CIFAR-100 & 23.64 & 0.59M & 132.7M
                             & Multi-stage pruning (300 epochs)   \\ 
    \hline
    ResNet-18 ($G=8$)        & ImageNet  & 44.71 & 1.91M & 0.20G
                             & Pruning and fine-tuning (30 epochs) \\ 
    ResNet-18 ($G=8$)~\cite{Zhao2018a}
                             & ImageNet  & 44.60 & 1.91M & 0.33G
                             & Training from scratch (90 epochs)\\ 
    \hline
  \end{tabular}
\end{table}

The key advantage of our method is achieving high accuracy under given inference budget constraints while still using a small amount of training budget.
To quantify this benefit and explain the final argument in the CIFAR results paragraph, we show the comparison between prior papers and ours on the amount of training effort to achieve similar model accuracy and inference budget.
Table~\ref{table:train_budget} organises experiments conducted in the paper and adds the actual training budget for each entry.
For all these cases, we use at most $1/3$ of the training budget from corresponding counterparts.
And our performance is competitive for each model.

\begin{figure}[ht]
  \centering
  \includegraphics[width=0.5\textwidth]{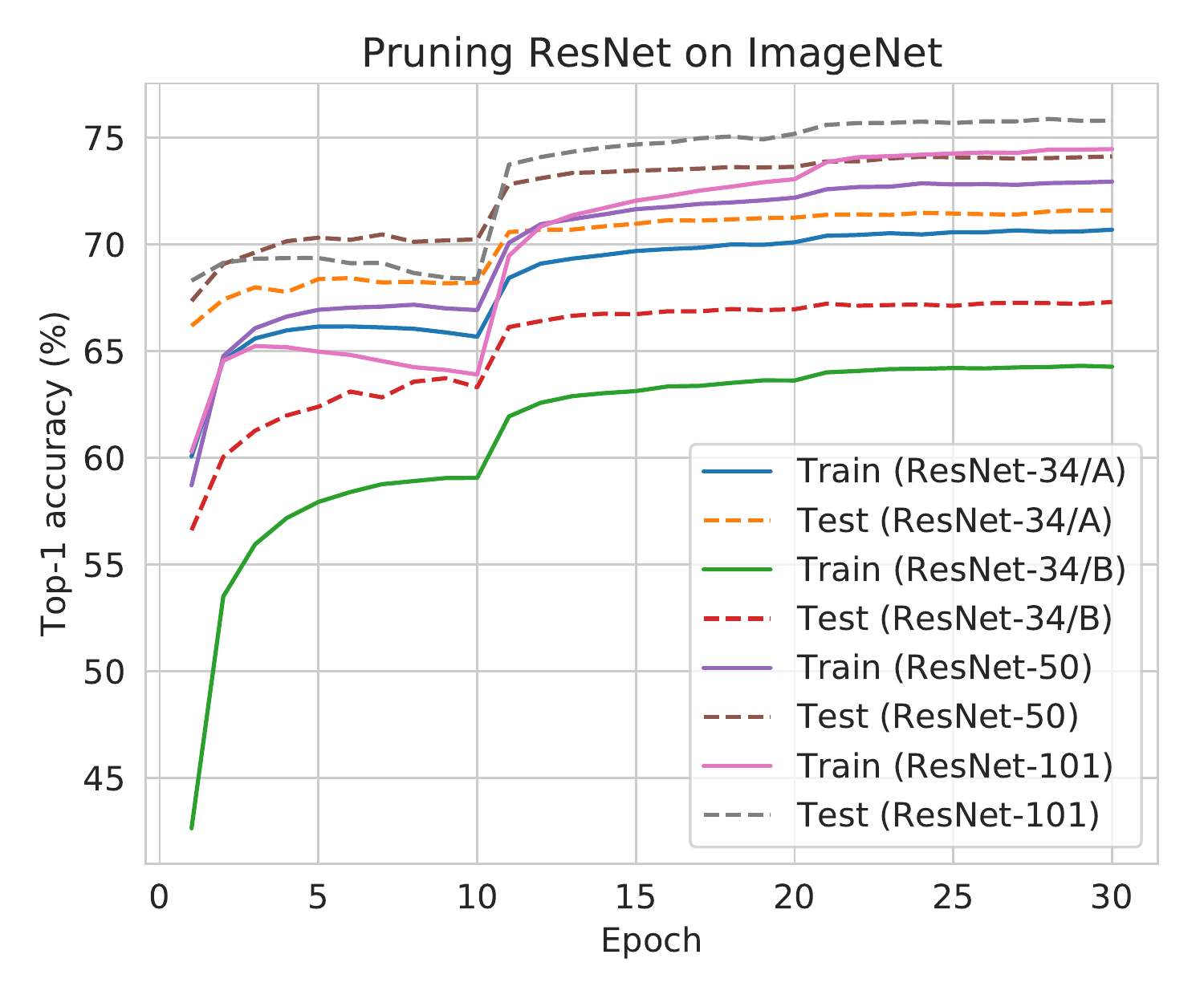}
  \caption{Fine-tuning ResNet pruning result on ImageNet.}
  \label{fig:imagenet}
\end{figure}

\paragraph{ImageNet pruning.}
\figref{fig:imagenet} illustrates the training curves of fine-tuning ResNet on ImageNet.
Our training strategy is simply taking $1/3$ of the original training schedule for the dense model.
It works well for ResNet-34/B which has most of ResNet-34's parameters removed, but for others the starting learning rate is slightly higher than required.
It is possible for us to achieve higher test accuracy if we can select a better set of hyperparameters.

\subsubsection{Comparison in Test Error with Prior Works}

\paragraph{CondenseNet comparison.}
In Table~\ref{table:train_budget}, we show that our pruned model has higher top-1 error than the one produced by~\cite{Huang2017a} at the scale of 2\% (DenseNet-86 is denoted as CondenseNet-86 in their paper and repository)\footnote{We try to reproduce their results with their open-source codebase using the same set of hyperparameters, while we can only achieve 24.68\% top-1 accuracy, which is about 1\% lower than ours.}.
There are two major reasons that are not demonstrated clearly in the paper:

\begin{enumerate}[itemsep=0em]
  \item Their approach can be considered as a training from scratch approach since they start from randomly initialised weights. And all $3\times 3$ convolution layers are initialised as group convolution and not produced by pruning. While in our case, we are restricted to prune weights and fine-tune in one-shot from a complete DenseNet-86 model, which significantly increases the difficulty of getting higher accuracy.
  \item Before they prune $1\times 1$ convolution into GConv, weights are being regularised by group-LASSO which can also make pruning easier. 
\end{enumerate}

In the future, we will look at how our method performs on a $1\times 1$ GConv only, which controls the factor from $3\times 3$ GConv layers trained from scratch.

\paragraph{Comparison with low-rank approximation.}

In our paper, we compare ResNet-34 pruning with~\cite{Peng2018}, and show that their accuracy is higher but the inference budget is increased as well.
Here we analysis the trade-off that is involved when comparing both approaches.

In terms of the inference budget, suppose $\CI=\CO=C$, the computation complexity of their pruned convolution layer will be $\OO(C^2\KH\KW/\GG + C^2)$, while in our case, the complexity is $\OO(C^2\KH\KW/\GG)$.
Therefore, they always add an extra amount of complexity $\OO(C^2)$.
Meanwhile, if $\KH=\KW=1$, their pruned complexity becomes $\OO(C^2/\GG+C^2)$, and since the original workload is $\OO(C^2)$, their result will increase the complexity instead of reducing them.
It further means that applying their method on $1\times 1$ convolution has negative effect and should be avoided.

On the other hand, regarding the training budget, although they use small amount of epochs for fine-tuning, the reconstruction time should be considered as well.
In their context, reconstruction means finding a matrix to multiply their approximation result that can minimise the different between the original output tensors and those produced by the pruned layer.
The formulation of this problem is listed as \eqref{eq:rec}, in which $Y$ is the original output, $Y^*$ is the approximated one, and $A^*$ is the matrix that can maximally reconstruct the output.
Even though this is a simple convex optimisation problem, solving it could be very time consuming.
Suppose $Y$ is a set of output tensors from a certain layer that is collected from all samples from the training dataset, and so for $Y^*$, then we can formulate $Y$ as a matrix of shape $(N, HW)$ where $N$ is the number of training samples and $HW$ is the number of pixels per image.
For ImageNet, $N$ is over 1.2 million, which indicates that this optimisation is almost intractable to solve.
We reckon they are using some advanced techniques to solve this problem, but still the whole training dataset should be iterated, while using our method does not need to do so and the complexity is only bounded by the number of parameters.
Therefore, the training budget of our method is at least the same as theirs.

\begin{equation}\label{eq:rec}
  A = \argmin_{A^*} \norm{Y - Y^*\times A^*}
\end{equation}

Their method indeed shows interesting insight into this GConv pruning problem and gives a closed-form solution, but there are some practical issues when applying them to models that consist of many $1\times 1$ convolution layers awaiting to be pruned or to large datasets.

\subsubsection{Fine-grained Recognition}

As mentioned before, this fine-grained recognition experiment only shows tentative results.
We still need to tune the hyperparameters to get a baseline model with better performance, and then try to prune the model that contains more informative weights.
Also, some recent fine-grained recognition papers intend to use more complicated network topologies that can learn discriminative features to distinguish samples with minor differences~\cite{Wang2016b}, which are better baseline models for exploration.
The difficulty to learn discriminative features is one of the reasons that requires us to fine-tune the pruned model with the same workload as the original training procedure.

\subsubsection{Model Exploration}

\begin{figure}[ht]
  \centering
  \includegraphics[width=\textwidth]{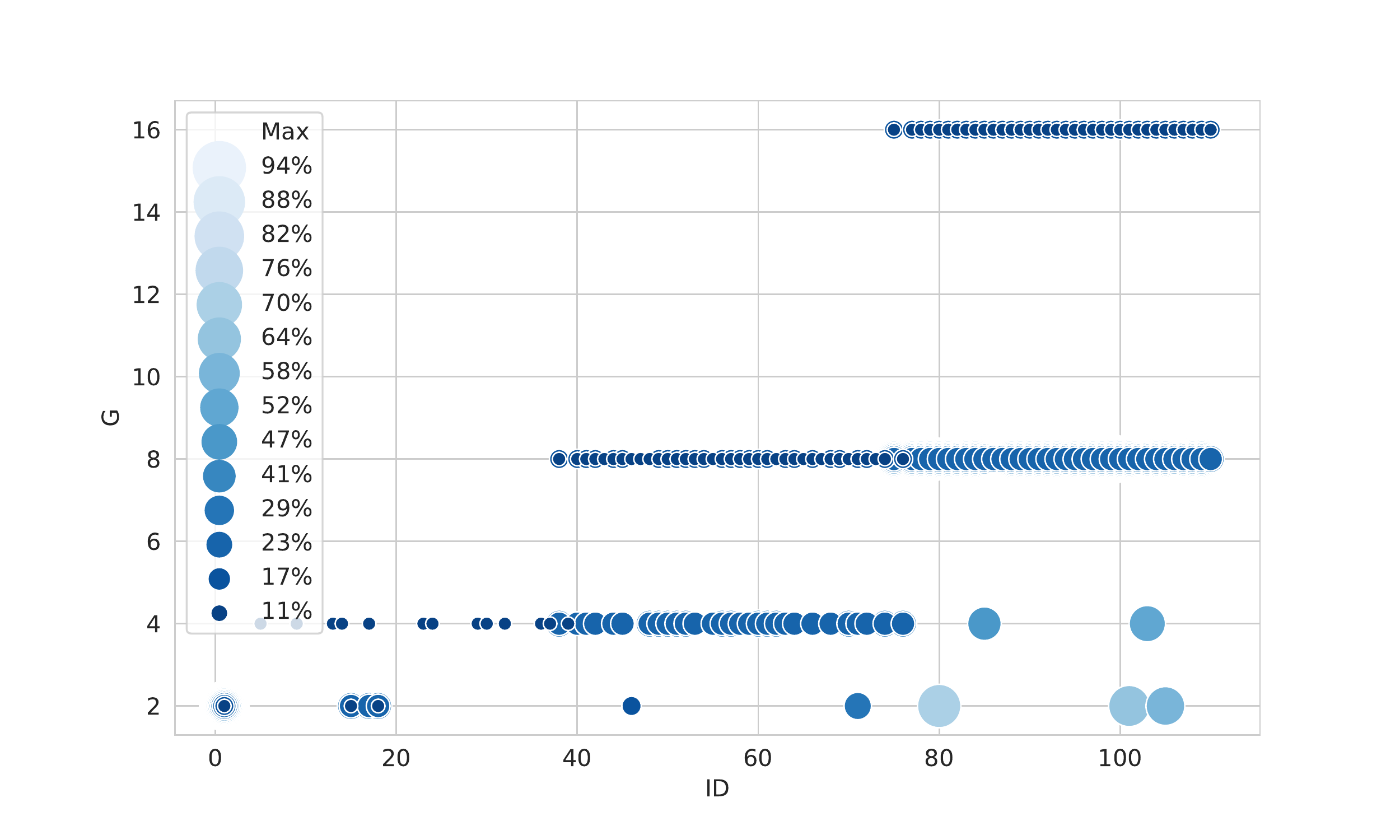}
  \caption{Illustrating the sparsity configuration (denoted by G) of each layer (denoted by ID) at different maximal number of parameters level (denoted by Max).}
  \label{fig:explore}
\end{figure}

Most of our exploration results are already listed in the paper.
Here we intend to summarise the structure of the models explored at different budget levels.
We visualise the distribution of $G$ among all layers for different budget in \figref{fig:explore}.
This figure shows that, with higher budget, layers with larger ID (closer to output) are more likely to be set with higher $G$ than other layers.
These sparsity configurations will be released with the codebase in the future.

\begin{figure}[ht]
  \centering
  \includegraphics[width=0.8\textwidth]{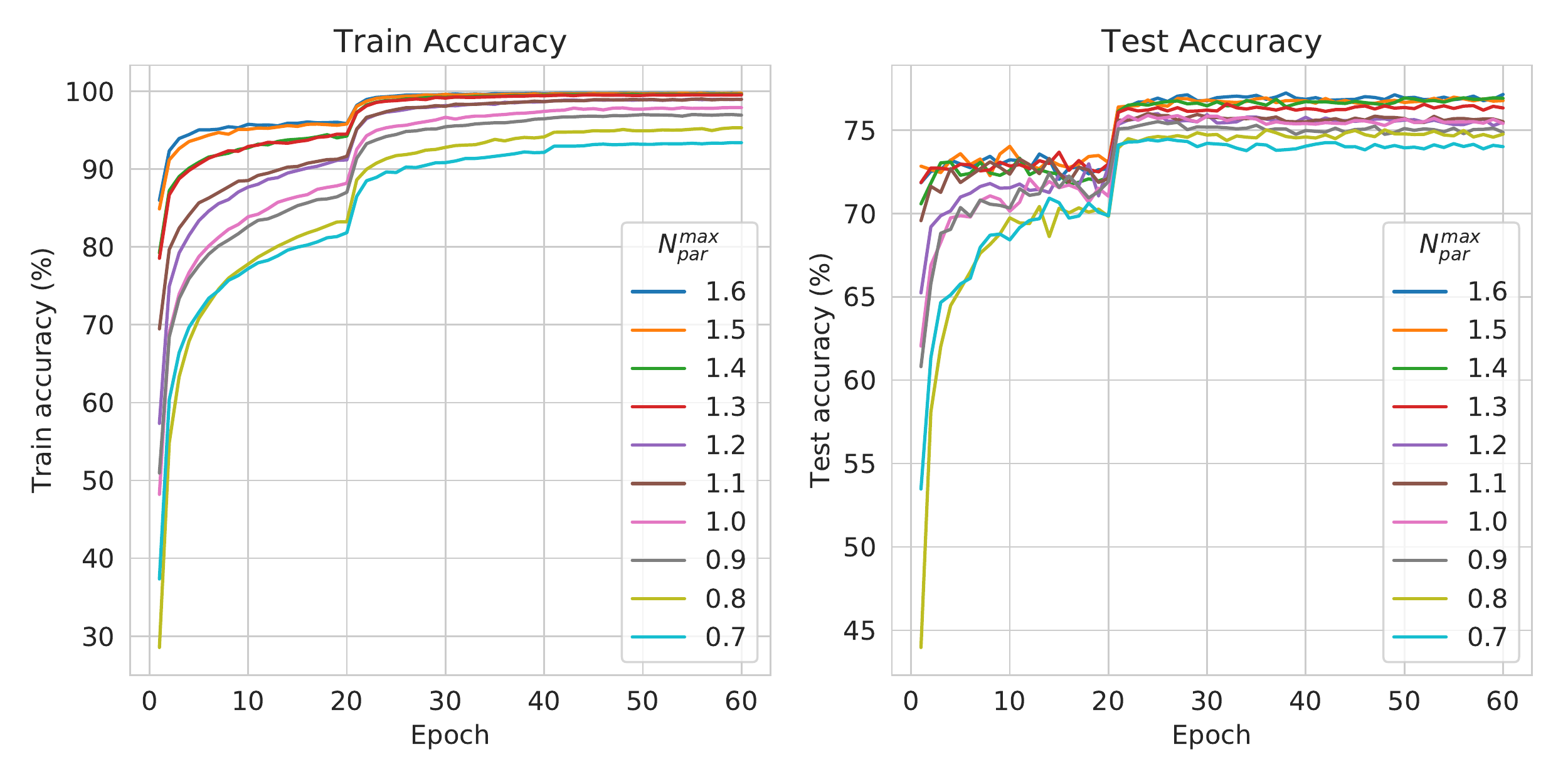}
  \caption{The curves of accuracy while fine-tuning those explored sparse models. Learning rate changes at 20 and 40 epochs.}
  \label{fig:explore:curves}
\end{figure}

We also list the training curves in~\ref{fig:explore:curves}. We notice that when the budget becoming more and more restrictive, it is more and more difficult to well train the model within a limited amount of epochs, e.g., when $\Nparmax<1.0$, the training accuracy is still increasing before the points that adjust the learning rate.

\subsubsection{Ablation Study}

\begin{figure}[ht]
  \centering
  \includegraphics[width=0.5\textwidth]{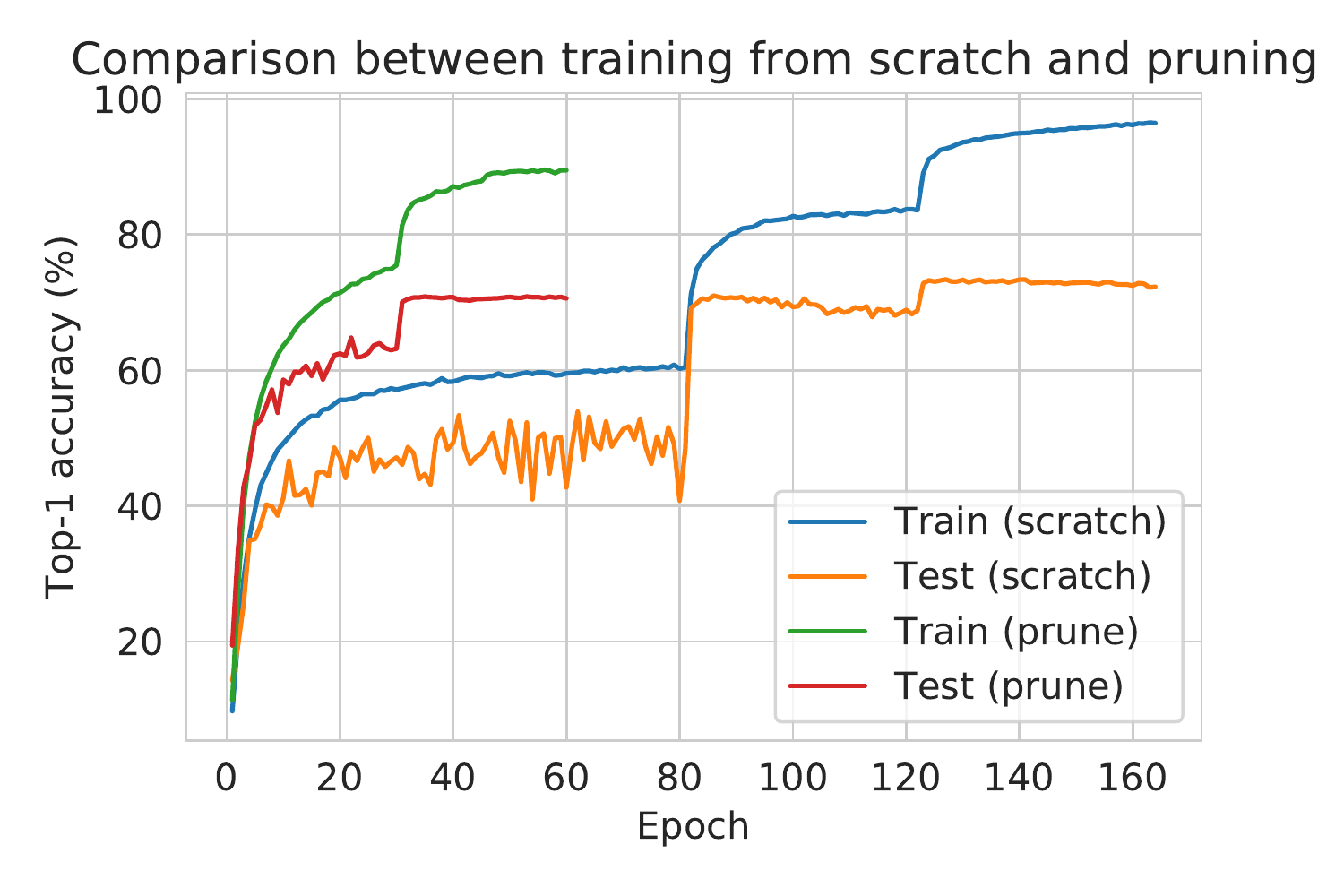}
  \caption{Curves for training from scratch and pruning by our method for ResNet-164 on CIFAR-100 with $G=4$.}
  \label{fig:ablation_scratch:curves}
\end{figure}

\paragraph{Compared to training from scratch.}
In our paper we mention that the main reason that our pruned model performs worse than the counterpart that is trained from scratch is due to the limitation on training time.
\figref{fig:ablation_scratch:curves} visualises this circumstance that the training accuracy of the model is still improving.

\subsubsection{Inference Speed}

\begin{figure}[ht]
  \centering
  \includegraphics[width=0.5\textwidth]{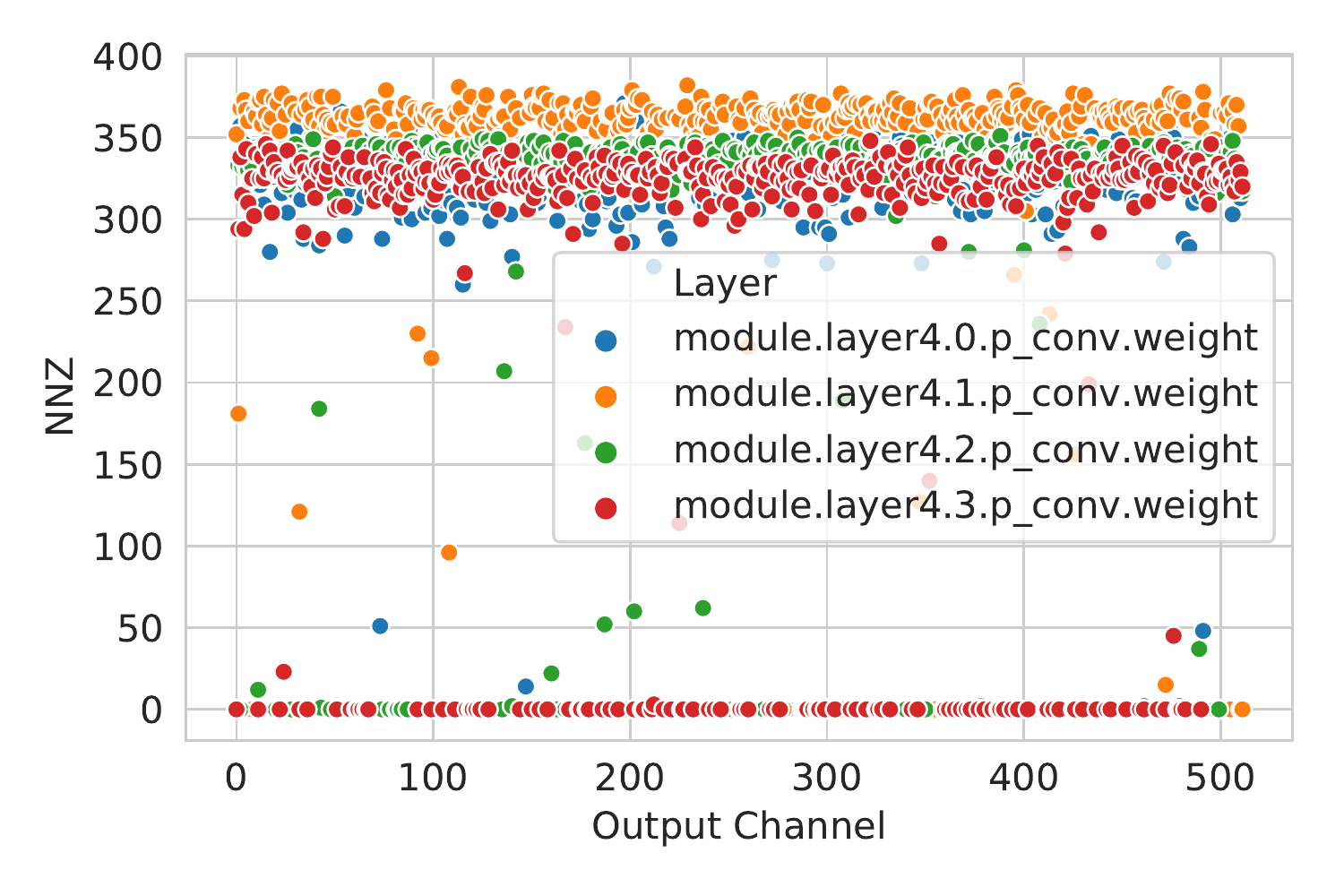}
  \caption{The distribution of non-zero values among $1\times 1$ sparse convolution layers in MobileNet. Here we select 4 layers with $\CO=512$ and $\CI=512$.}
  \label{fig:sparse}
\end{figure}

In the paper we show that our resulting MobileNet is faster than an irregular sparse MobileNet model, pruned by~\cite{Zhu2017}.
One major reason is that the performance of sparse matrix multiplication (SpMM), the primitive we use to implement sparse $1\times 1$ convolution, highly depends on the sparsity pattern.
Weights is a sparse matrix in this case, and if it is irregular sparse, then we cannot efficiently map the matrix multiplication workload into multiple parallel SIMD primitives without overhead, so that the performance will be degraded.
It is obvious that all the sparse weights in our pruned models are perfectly regular sparse, i.e., for each row (output channel) the number of non-zero kernels stays the same.
We illustrate the sparsity pattern of MobileNet in \figref{fig:sparse} as well, and we can notice that across output channels, the distribution of non-zero values are indeed irregular, which is harmful to performance.

\newpage

{\small
\bibliographystyle{ieee}
\bibliography{library,custom}
}